\documentclass[sigconf,9pt]{acmart}
\AtBeginDocument{%
  }

\usepackage{threeparttable}
\usepackage{latexsym}
\usepackage{graphicx}
\usepackage{diagbox} 
\usepackage{amsmath}
\usepackage{multirow}
\usepackage[table]{xcolor}  

\usepackage{booktabs}
\usepackage{comment}
\usepackage[linesnumbered,ruled,vlined]{algorithm2e}
\usepackage{enumitem}
\usepackage{xspace}
\usepackage{makecell}
\usepackage{newtxmath}
\usepackage{placeins}
\usepackage{booktabs} 
\usepackage{array}    

\usepackage{xspace}
\usepackage{tablefootnote}
\usepackage{threeparttable}
\usepackage{comment}
\usepackage{graphicx}
\usepackage{amsmath}
\usepackage{multirow}

\usepackage{newtxmath}
\usepackage{placeins}
\usepackage{booktabs} 
\usepackage{array}    
\usepackage{threeparttable}
\usepackage{pifont}
\usepackage{subcaption}
\usepackage{balance} 
\newcommand{\circled}[1]{\ding{\numexpr171+#1\relax}}

\newcommand{\compactparagraph}[1]{\noindent{\textbf{\textit{#1}}}}
\newcommand{\projectnamenott}{FBLayout}
\newcommand{\projectname}{{\tt \projectnamenott}\xspace}

\long\def\wniu#1{\textcolor{brown}{WN: #1}}
\usepackage{url}

\copyrightyear{2026}
\acmYear{2026}
\setcopyright{cc}
\setcctype{by-nc-nd}
\acmConference[MobiSys '26]{The 24th Annual International Conference on Mobile Systems, Applications and Services}{June 21--25, 2026}{Cambridge, United Kingdom}
\acmBooktitle{The 24th Annual International Conference on Mobile Systems, Applications and Services (MobiSys '26), June 21--25, 2026, Cambridge, United Kingdom}
\acmDOI{10.1145/3745756.3809214}
\acmISBN{979-8-4007-2027-7/2026/06}

\begin{document}

\title{FBLayout: Optimizing Memory Layout for Efficient LLM Finetuning on Mobile GPUs}


\author{Kahou Tam}
\orcid{0000-0001-5816-6837}
\authornote{Equal Contribution.}
\affiliation{%
   \position{State Key Laboratory of IoTSC} 
  \institution{University of Macau}
  \state{Macau SAR}
  \country{China}
}
\email{tamkahou.2023@connect.um.edu.mo}

\author{Wei Niu}
\authornotemark[1]
\affiliation{%
  \institution{University of Georgia}
  \state{Athens, GA}
  \country{USA}}
\email{wniu@uga.edu}

\author{Yu Bao}
\authornotemark[1]
\affiliation{%
   \position{State Key Laboratory of IoTSC} 
  \institution{University of Macau}
  \state{Macau SAR}
  \country{China}
}
\email{mc45002@um.edu.mo}

\author{Xiaomin Ouyang}
\affiliation{%
  \institution{Hong Kong University of Science and Technology}
  \state{Hong Kong}
  \country{China}}
\email{xmouyang@cse.ust.hk}

\author{ChengZhong Xu}
\orcid{0000-0001-9480-0356}
\affiliation{%
\position{State Key Laboratory of IoTSC} 
  \institution{University of Macau}
  \state{Macau SAR}
  \country{China}}
\email{czxu@um.edu.mo}

\author{Li Li}
\authornote{Corresponding author}
\orcid{0000-0002-2044-8289}
\affiliation{%
\position{State Key Laboratory of IoTSC} 
  \institution{University of Macau}
  \state{Macau SAR}
  \country{China}}
\email{llili@um.edu.mo}

\begin{abstract}
Transformer-based models have enabled unprecedented capabilities across language, vision, and multimodal tasks. On-device fine-tuning of transformer models offers a privacy-preserving path to personalized AI, yet remains inefficient on mobile GPUs due to severe memory constraints and frequent layout transformations in attention mechanism during training. 
Existing mobile training frameworks either use unified layouts for forward and backward passes -- leading to fragmented memory access and poor GPU utilization during backpropagation -- or rely on explicit layout conversions, which introduce significant transformation overhead.

To overcome this, we propose \projectname, a layout-aware framework that co-designs tensor organization with mobile GPU platforms. \projectname introduces: (1) a unified R-Tile layout for multi-dimensional reductions across forward/backward passes; (2) tile-based index transformation to eliminate physical data movement; and (3) activation-guided layout selection to propagate efficient layouts globally. Evaluations on seven transformer models across different mobile phones (including ARM Mali and Qualcomm Adreno GPUs) show that \projectname achieves 2.2-5.7$\times$ speedup over MNN, TFLite, and TVM, while significantly improving cache efficiency and reducing memory footprint, enabling practical on-device large model fine-tuning.

\end{abstract}

\ccsdesc[300]{Computer systems organization~Embedded systems}
\ccsdesc[300]{Computing methodologies~Artificial intelligence}

\keywords{On Device Training, LLM Fine-tuning, Mobile GPUs Optimization.}




\maketitle

\section{Introduction}
The transformative power of transformer-based models in language, vision, and multimodal tasks has driven a new wave of intelligent mobile applications~\cite{li2024large,xu2024device,tian2025clone,tian2026floe}. To deliver personalized experiences while addressing data privacy concerns, it has become essential to adapt models directly on ubiquitous devices~\cite{tseng2024two,zhang2024personalization,tam2023federated,tam2023fedcoop}. This need arises across diverse scenarios: personalized LLM agents that continuously adapt to evolving user preferences and language styles \cite{park2023generative,xu2024survey}, privacy-preserving recommendation systems where sensitive behavioral data must remain on-device \cite{zhao2024recommender,lin2025can,tsai2024leveraging,wu2025survey,tian2022harmony}, and test-time adaptation where deployed models face real-world distribution shifts (e.g., speech models encountering unseen accents) requiring immediate local updates \cite{hu2025test,karmanov2024efficient}. In these settings, offloading to cloud or on-premises servers is often infeasible due to latency, connectivity, privacy, or regulatory constraints. On-device fine-tuning, which updates pre-trained models with local sensitive data, thus becomes a promising approach to enable customized model behavior without exposing private information.

However, efficient on-device fine-tuning is significantly limited by mobile resource constraints.
The challenges are threefold: {\it high memory usage}, {\it intensive computation demands}, and {\it excessive data layout transformations}.
First, training requires storing intermediate activations for gradient computation, leading to memory footprints 7–10$\times$ larger than inference \cite{devlin2019bert}, often exceeding the typical memory capacity of modern devices \cite{24Gphone}. 
While techniques like LoRA \cite{hu2022lora} and activation recomputation \cite{gim2022memory,wang2022melon} alleviate memory pressure, they remain largely ineffective against the computational inefficiencies induced by frequent data layout transformations on mobile platforms. 
Second, backward propagation is also far more computationally demanding. 
Fine-tuning even a modest 1B-parameter model on a mobile CPU can take over 50 seconds per step~\cite{li2025mobillm}. With practical fine-tuning often requiring thousands of steps, users may wait hours to days before the model reflects their latest data and preferences, significantly degrading the responsiveness of on-device personalization.  Moreover, prolonged training induces thermal throttling on mobile devices, progressively degrading throughput and potentially preventing training from completing~\cite{dettmers2023qlora,wang2022melon,wu2024bridging,wu2025memory,wu2025breaking,wu2025elastic,wu2026beyond}. Reducing fine-tuning latency is therefore essential to making on-device model adaptation practical and user-friendly.
Third, and most critically, transformer fine-tuning involves frequent data layout transformations (e.g., reshapes, transposes)  inherent to attention and matrix operations \cite{ainslie2023gqa}, which disrupt data locality and strain mobile memory bandwidth \cite{niu2024smartmem}.

While significant research has optimized on-device inference \cite{jiang2020mnn,chen2018tvm,niu2021dnnfusion,liang2022romou}, 
most frameworks \cite{jiang2020mnn,TFlite} primarily focus on forward-pass optimizations and overlook the bidirectional data flow required for training. 
Existing mobile training frameworks employ suboptimal strategies to manage the resulting forward-backward layout conflict: 
some reuse a single layout for both passes, leading to inefficient access patterns during backward operations \cite{jiang2020mnn}; 
others introduce explicit layout conversions, resulting in considerable overhead and memory fragmentation \cite{niu2024smartmem}. 
Both approaches fail to address the fundamental conflict and, importantly, do not co-design layouts with the mobile GPU's texture memory architecture \cite{guan2025tmmodel,hakura1997design}, 
which is optimized for 2D spatial locality, thus leaving significant performance on the table.

To tackle these limitations, we introduce a layout-aware framework, 
\projectname (\textbf{F}orward-\textbf{B}ackward unified \textbf{Layout}), 
that incorporates a list of novel techniques to optimize tensor layouts 
for efficient transformer fine-tuning on mobile GPUs.
Specifically, we propose:

$\bullet$  A high-level classification of operator types that categorizes them based on their sensitivity to forward-backward dependencies and memory access consistency.

$\bullet$ A unified, texture-aware R-Tile layout that resolves conflicting access patterns between forward and backward passes, allowing for efficient multi-dimensional access without explicit transformations.

$\bullet$  An R-Tile-based index transformation procedure that eliminates expensive layout operations (e.g., reshape and transpose) by replacing physical data movement with lightweight coordinate remapping.

$\bullet$  An activation-guided global layout selection strategy that propagates and refines layout decisions across the entire computation graph.

To the best of our knowledge, \projectname is the \textbf{first} work to accelerate LLM fine-tuning on mobile GPUs. 
We extensively evaluate \projectname on 7 representative Transformer-based models spanning decoder, encoder, and encoder-decoder architectures, deployed on two mobile GPU platforms (ARM Mali and Qualcomm Adreno). 
Compared against three state-of-the-art frameworks (MNN \cite{jiang2020mnn}, TVM \cite{chen2018tvm}, TFLite \cite{TFlite}), \projectname consistently achieves superior performance: it delivers speedups of up to 4$\times$, 5.8$\times$, and 5.8$\times$, respectively. 
These gains are especially pronounced in LLMs and diffusion-based workloads, 
where complex data reorganization dominates cost. 
Furthermore, \projectname significantly improves cache efficiency and reduces memory pressure, 
enabling stable and efficient fine-tuning under tight memory constraints compared to existing frameworks.

\section{Background and Motivation}

\begin{figure}[!t]
  \centering
  \includegraphics[width=0.98\linewidth]{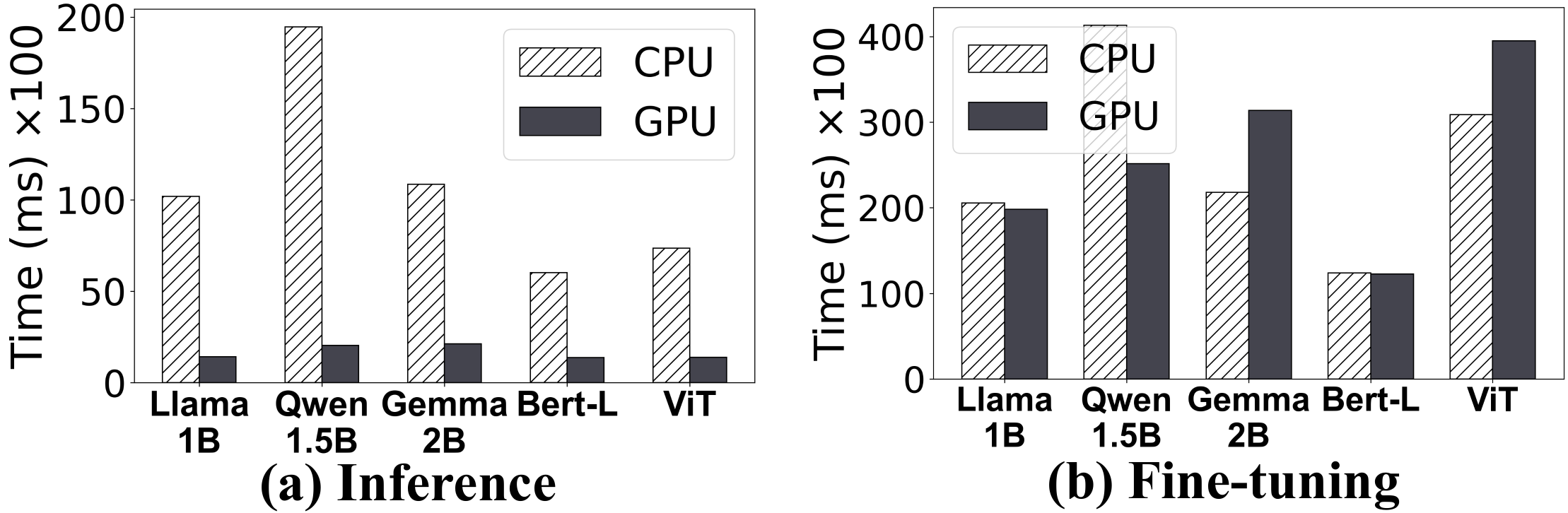}
  \vspace{-5pt}
  \caption{The latency comparison of on-device LLM inference and training on mobile (Snapdragon 8 Elite).}
  \label{inferencevsTrainintime}
\end{figure}

\subsection{Performance Gap in LLM Fine-Tuning}
The need for on-device LLM fine-tuning is evident \cite{tseng2024two,zhang2024personalization,tam2024fedhybrid,lin2025can}; however, 
the fine-tuning process places significantly greater and more complex demands on device resources. 
This creates unique system-level challenges that current inference-oriented frameworks struggle to address.
The transition from inference to fine-tuning on a mobile SoC highlights a significant performance gap. 
While mobile GPUs typically offer substantial speedups (approximately 6.5 $\times$ in Figure~\ref{inferencevsTrainintime}) for inference compared to CPUs, this advantage diminishes during fine-tuning. 
As illustrated in Figure~\ref{inferencevsTrainintime}, the fine-tuning latency on GPUs can be comparable to, or even worse than (by 43\%), 
that on CPUs for models with 0.5B to 2B parameters. 
The performance gap primarily stems from the increased complexity of fine-tuning workloads, 
which include not only forward passes but also backward passes and weight updates. 

\subsection{Root Cause: FB-Layout Conflict}

The key inefficiency arises not only from limited compute capability, 
but from a fundamental mismatch between the bidirectional dataflow of training and 
the memory hierarchy on mobile GPUs. 
Unlike inference, training requires tensors (e.g., weights, intermediate activations, and states) 
produced during the forward pass to be reused in the backward pass for gradient computation. 
Due to the mathematics of backpropagation, however, the same tensor is often accessed with 
different and even conflicting patterns in the two passes—typically involving transpositions 
and reductions along different dimensions.

\begin{figure}[!t]
  \centering
  \includegraphics[width=0.97\linewidth]{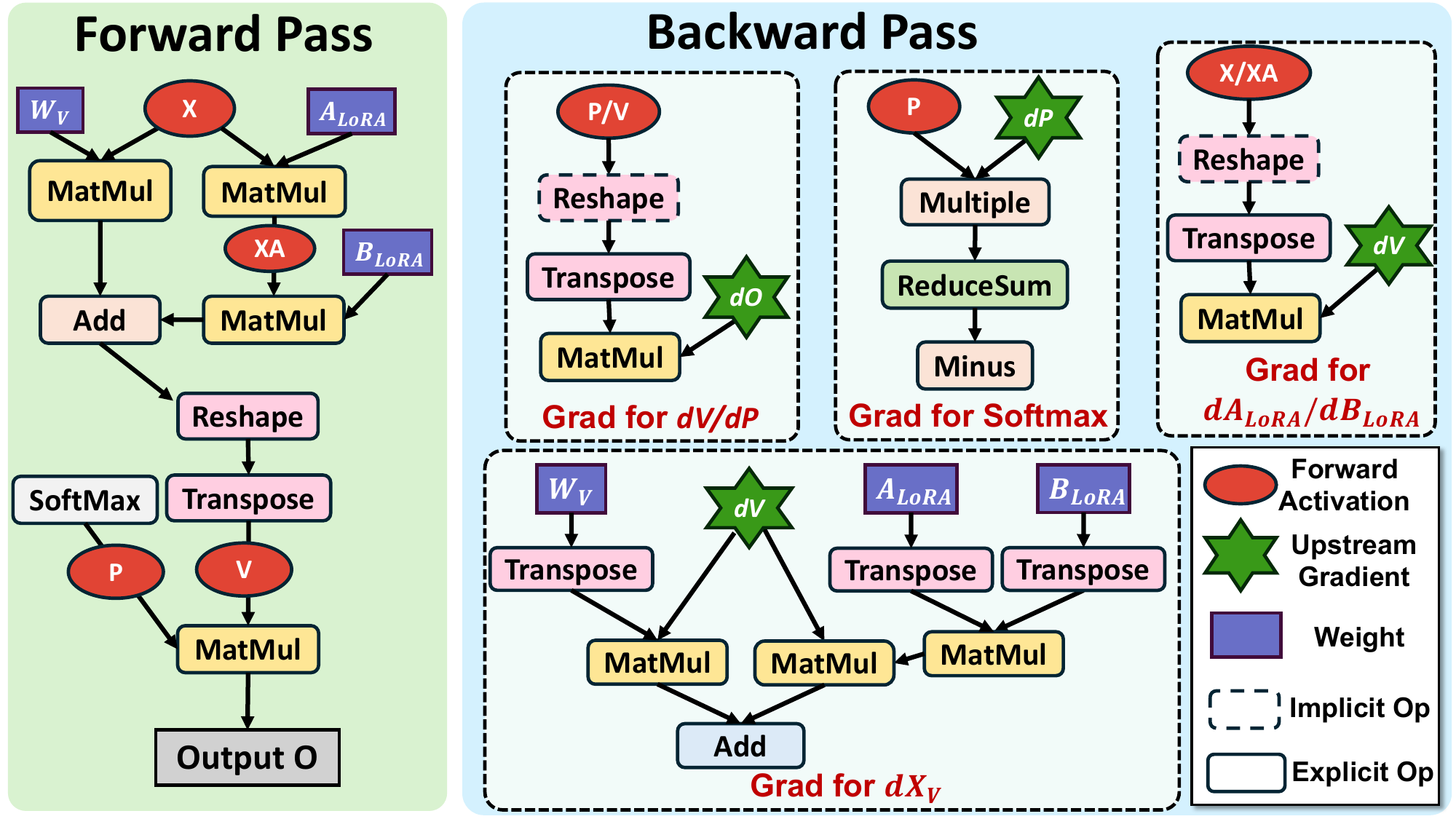}
  
  \caption{
  Example of the forward-backward layout conflict in LoRA fine-tuning of the attention module. LoRA~\cite{hu2022lora} injects low-rank adapter matrices $A_{LoRA}$ and $B_{LoRA}$ to approximate weight updates, avoiding full-parameter training. Here $X$ is the input, $XA$ the intermediate activation from the LoRA path, $P = \text{softmax}(QK^\top)$ the attention scores, and $V$ the value tensor. The backward pass introduces transpose and reshape operators that conflict with the forward layout.}
  \label{MotivationExample}
  \vspace{-10pt}
\end{figure}

Figure~\ref{MotivationExample} illustrates the computational graph of forward and backward passes in the attention module with LoRA fine-tuning.  
Compared to inference workloads, training introduces two critical characteristics at the workload level. 
{\em First, training creates complex tensor dependencies}: unlike inference where each tensor is typically consumed once, training requires forward-pass activations (e.g., $X$, $XA$) and weights (e.g., $A_{LoRA}$, $B_{LoRA}$, $W_V$) to be reused during backward computation for gradient calculation, resulting in multiple consumers per tensor and intricate data flow patterns. {\em Second, backpropagation demands extensive data reorganization operations}: transpose operations (e.g., $X^T$, $A_{LoRA}^T$, $B_{LoRA}^T$, $W_V^T$) are mathematically inherent to gradient computation via the chain rule, while reshape operations arise both explicitly and implicitly when tensors must be reorganized to match dimensional requirements across forward and backward passes, as weight matrices and activations often participate in computations with different dimensional configurations. 
It is worth noting that,  this issue is particularly pronounced in transformer-based models, as diverse attention mechanisms (e.g., GQA \cite{ainslie2023gqa}, MHA \cite{vaswani2017attention}, cross-attention \cite{chen2021crossvit}) require unique and incompatible approaches to tensor partitioning and layout.

\begin{figure}[!t]
  \centering
  \includegraphics[width=0.99\linewidth]{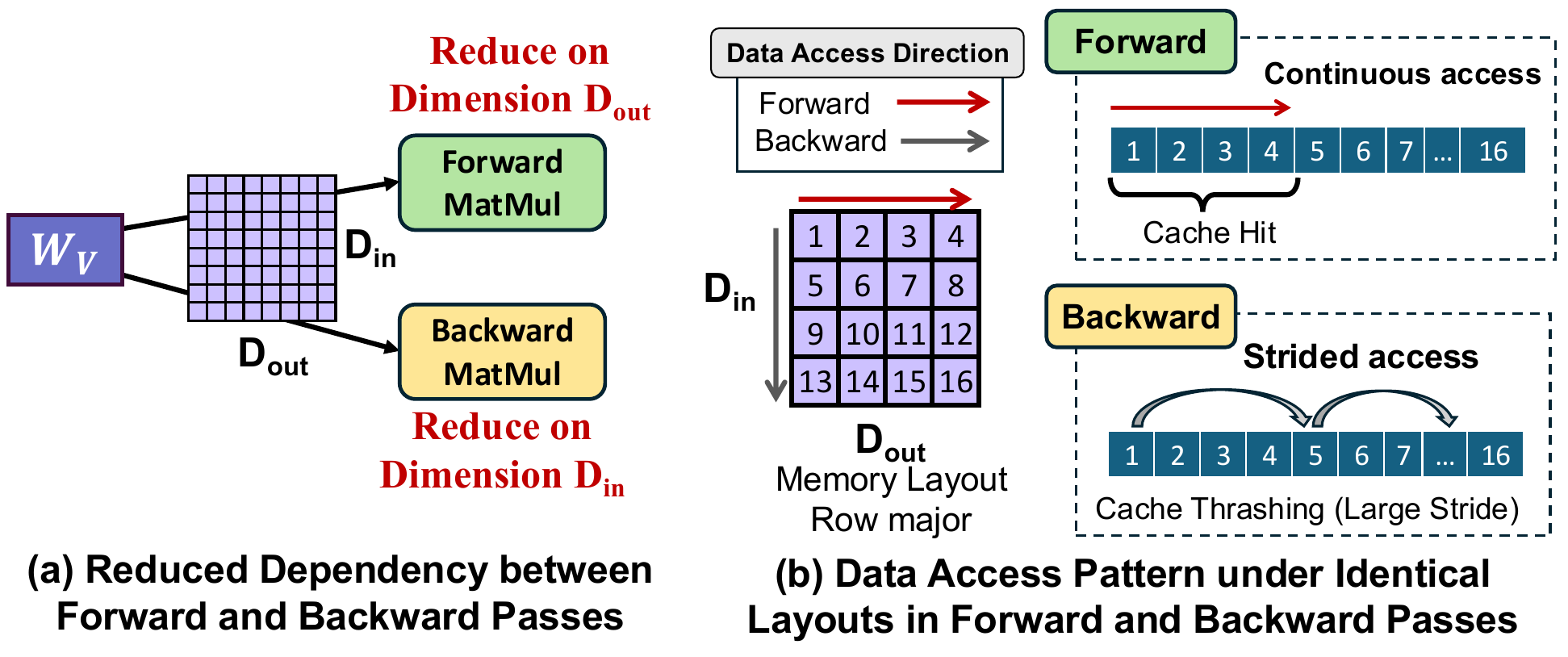}
  \vspace{-5pt}
  \caption{Example of FB-layout conflict for a weight matrix. 
(a) Forward and backward MatMul reduce along different dimensions; 
(b) resulting contiguous (forward) vs. strided (backward) accesses when sharing one layout in  memory.}
  \label{Reduceexample}
  \vspace{-10pt}
\end{figure}

This creates a forward-backward layout (FB-Layout) conflict. 
Figure~\ref{Reduceexample} illustrates this using a weight matrix ($W_V$) from Figure~\ref{MotivationExample}. 
A layout optimized for spatial locality in the forward pass forces the backward pass into strided, 
non-contiguous memory accesses. 
This conflict is further intensified by the distinct memory hierarchy of mobile GPUs. Mobile GPU caches are limited \cite{guan2025tmmodel}, and strided access patterns from layout conflicts cause frequent cache misses. Each miss requires fetching data from DRAM, where mobile low memory bandwidth ($<100$ GB/s) \cite{adreno_gpu} leads to severe latency penalties.

\begin{figure}[!t]
  \centering
  \includegraphics[width=0.9\linewidth]{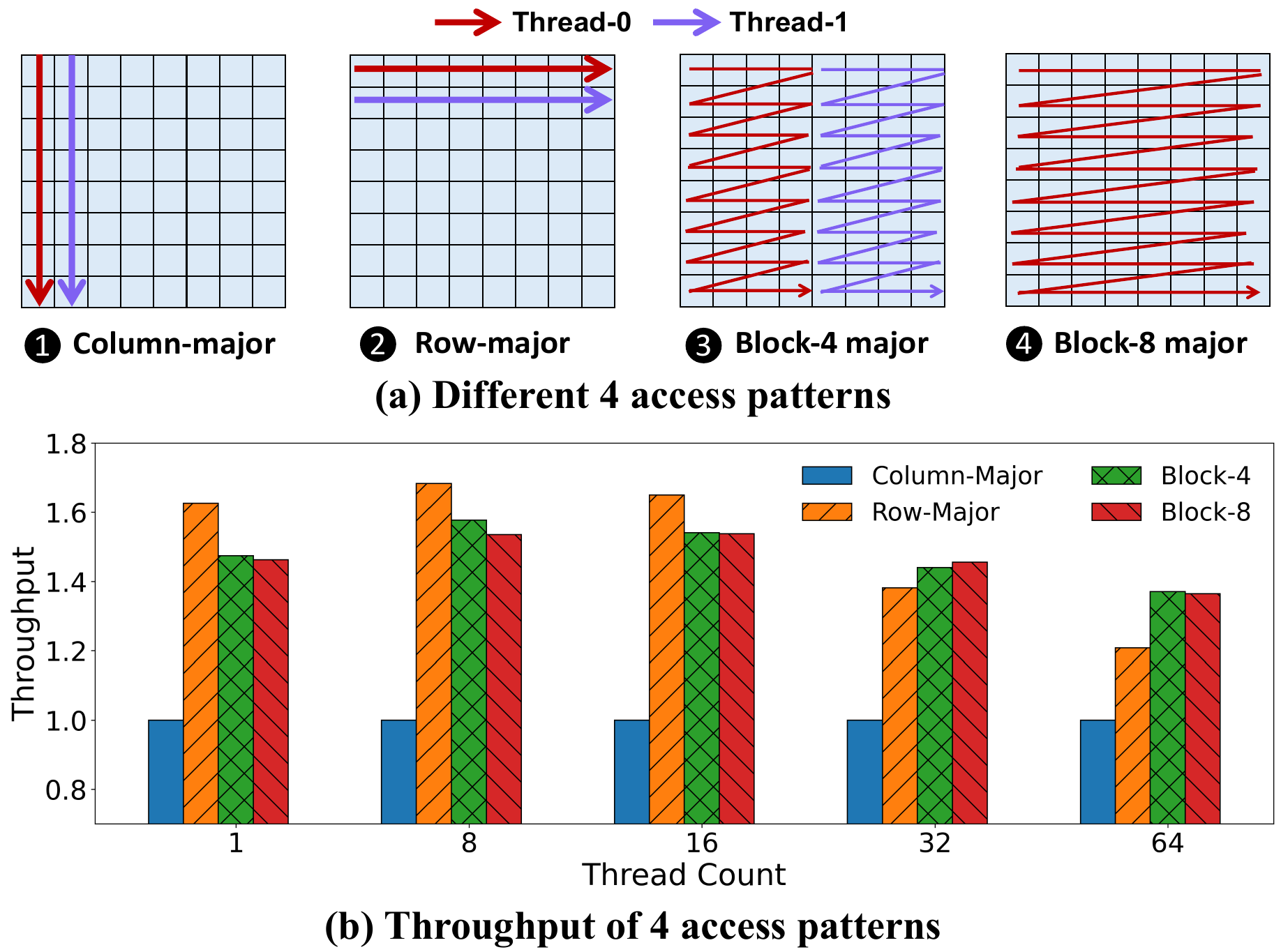}
  \vspace{-5pt}
  \caption{Texture memory spatial locality. (a) Different 4 access patterns; (b) Throughput of different 4 access patterns. Experimental settings detailed in supplementary material.}
  \label{2DSpatiallocality}
  \vspace{-10pt}
\end{figure}

Mobile GPUs employ a 2.5D texture memory system \cite{hakura1997design}, 
where tensors are mapped to a 2D coordinate space ({\em width}, {\em height}); 
each coordinate stores a 4-element RGBA vector. The accompanying texture cache 
is explicitly optimized for 2D spatial locality \cite{tmu}, 
meaning accesses to adjacent coordinates are highly efficient. 
Consequently, access patterns that are contiguous in this 2D grid (e.g., blocked traversals) 
achieve significantly higher bandwidth than strided or scattered patterns, 
as quantified in Figure~\ref{2DSpatiallocality}. 
The forward-backward layout conflict directly undermines this locality: 
a layout favoring one pass typically violates the 2D contiguous access 
required for efficiency in the other pass. 
This leads to severe cache thrashing, 
underutilization of SIMD units (evidenced by low ALU utilization in Figure~\ref{timeanalysis}(b)), 
and pipeline stalls due to frequent accesses to slow global memory.

\subsection{Limitations of Existing Solutions.}

\begin{figure}[!t]
  \centering
  \includegraphics[width=0.98\linewidth]{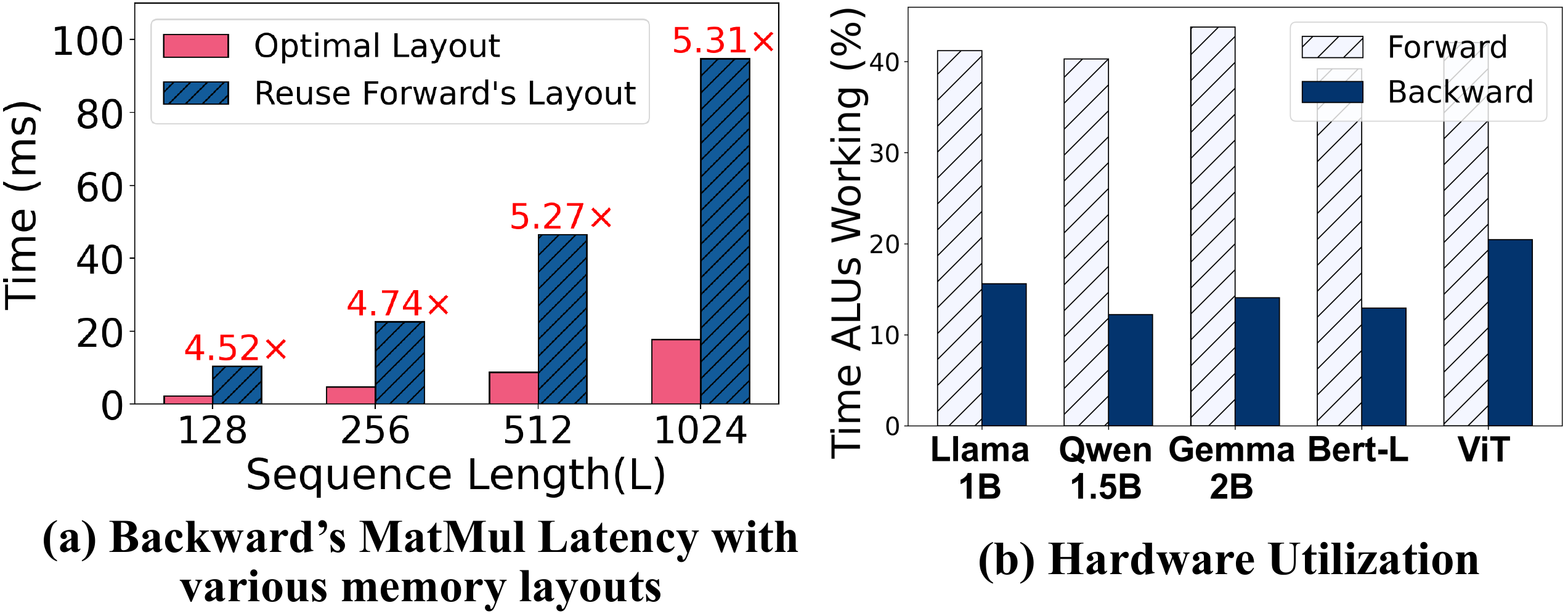}
  \vspace{-5pt}
  \caption{Performance overhead of reusing forward layouts in backward pass (MNN). (a) MatMul latency: optimal backward layout vs. reused forward layout ($D_{in}=D_{out}=2048$, varying sequence length $L$). (b) GPU ALU utilization degradation across LLM training models under layout reuse.}
  \label{timeanalysis}
\end{figure}

\begin{figure}[!t]
  \centering
  \includegraphics[width=0.98\linewidth]{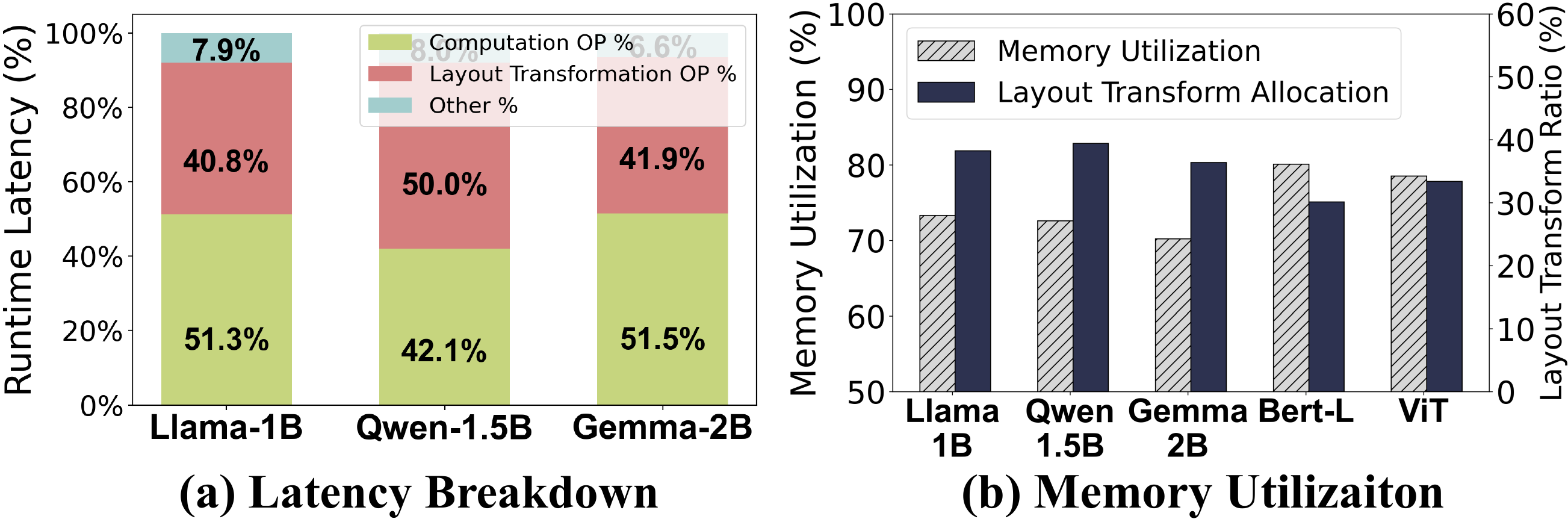}
  \vspace{-5pt}
  \caption{Overhead analysis of explicit layout transformations during LLM training: (a) latency distribution across operators; (b) memory consumption analysis.}
  \label{overheadeplicttransformation}
\end{figure}

To address this, existing mobile frameworks utilize two distinct, yet similarly inefficient, strategies to address this conflict (see Figures~\ref{timeanalysis} and \ref{overheadeplicttransformation}).
{\it Layout Reuse} (e.g., MNN \cite{jiang2020mnn}): 
A single layout from the forward pass is reused in backward, avoiding transformation overhead 
but forcing backward kernels into highly inefficient, non-contiguous memory accesses, 
causing up to 5.3$\times$ slowdown for key operations.
{\it Explicit Transformations} (e.g., TFLite \cite{TFlite}): 
Explicit layout conversion operators (e.g., {\em transpose}, {\em reshape}) are inserted 
to provide each kernel its optimal layout. This preserves per-kernel efficiency 
but introduces prohibitive overhead-consuming up to 50\% of total runtime 
and 30\% (or more) of memory allocations-while exacerbating memory fragmentation \cite{niu2024smartmem}.

Both strategies view the layout conflict as an unavoidable cost, leading to both inefficiency and sub-optimal performance. 
The key issue remains: there is no unified data layout designed in conjunction with the 
mobile GPU's texture memory hierarchy to effectively accommodate the access patterns 
of both forward and backward passes.  
It is worth noting that while mobile NPUs offer significantly higher computing capability, they lack backpropagation support, operate at limited precision (typically int8) \cite{xu2022mandheling}, and provide only fixed-architecture inference acceleration \cite{npu}. Thus, mobile GPUs remain the only practical on‑device accelerator, which makes our layout optimization challenge especially critical.


\section{Problem Formulation and Analysis}
\label{sec:formalization}

This section formalizes the layout optimization challenges for LLM fine-tuning on mobile GPUs. 
We start by modeling the computation graph and memory architecture constraints, 
then clearly define the key problems to be solved.

\vspace{-5pt}
\subsection{Computational Graph in LLM Training}

Let the training computation be represented as a directed acyclic graph $G = (V, E)$, where 
vertices ($V = V_{fwd} \cup V_{bwd}$) represent operators in forward and backward passes; 
edges ($E = E_{fwd} \cup E_{bwd} \cup E_{reuse}$) represent tensor dependencies.
Here $E_{reuse}$ denotes edges where forward tensors are reused in backward pass.
Each tensor $t \in E$ has logical dimensions $S(t) = (d_1, d_2, \dots, d_n)$ 
and requires a physical layout $L(t)$ for GPU memory storage. 
The key distinction from inference graphs is the bidirectional data flow: 
{\it backward operations $v \in V_{bwd}$ depend on forward activations through $E_{reuse}$}.

The mobile GPU memory architecture further complicates this bidirectional optimization. 
Tensors must be mapped to 2.5D texture memory using 2D spatial coordinates $(W, H)$ 
where each location stores a 4-element RGBA vector \cite{niu2024smartmem}. This coordinate-based indexing, combined with a 2D texture cache optimized for spatial locality \cite{hakura1997design}, creates unique constraints that fundamentally differentiate mobile GPU optimization from server-grade training systems \cite{paszke2019pytorch}.

This computational framework reveals several critical optimization challenges that 
existing mobile training solutions handle suboptimally. {\it The reduction-axis conflict problem} 
arises when forward and backward operations require incompatible data layouts from the same tensor. 
For instance, matrix multiplication reduces along different dimensions during forward propagation 
versus gradient computation, creating a fundamental tension where no single layout can be optimal 
for both passes. This conflict is particularly severe on mobile GPUs where the 2D texture cache 
heavily penalizes non-contiguous access patterns.

{\it The global layout selection problem} extends these local conflicts to the entire training graph, 
requiring coordinated layout decisions across all operators. The optimization objective is to assign 
layouts $L(t)$ for all tensors $t \in E$ to minimize total execution cost, balancing computational 
efficiency against transformation overhead. This represents a complex combinatorial optimization 
problem due to the interdependence of layout decisions and mobile memory constraints.

\subsection{Formal Problem Statements}

Based on the identified challenges, we formulate four key research problems that must be 
addressed to enable efficient LLM fine-tuning on mobile GPUs.

{\it $Q1$: Unified Layout Design:} For tensors with reduction layout conflicts, determine if there exists 
a unified layout $L_u(t)$ that provides near-optimal performance for both forward and backward passes 
while avoiding the cost of explicit transformations. The challenge lies in constructing such layouts 
that exploit mobile GPU texture memory characteristics despite conflicting access patterns.

{\it $Q2$: Transformation Elimination:} Under what conditions can sequences of transformation operations 
be replaced with logical coordinate mappings at negligible cost? This requires determining when 
physical data transformations can be collapsed into lightweight index computations while preserving 
semantic correctness and maintaining cache efficiency.

{\it $Q3$: Global Layout Optimization:} What practical algorithms can solve the global layout assignment 
problem with bounded suboptimality for real-world training graphs? The exponential search space 
necessitates efficient heuristics that consider tensor reuse patterns and fusion opportunities 
across the entire computation graph.

{\it $Q4$: 2.5D Memory Mapping:} How can we efficiently implement unified layouts on 2.5D texture memory 
while respecting coordinate-based indexing and RGBA channel organization? This requires developing 
cache-aware mapping strategies that maintain efficiency across diverse access patterns inherent 
in training workloads.

These formalized problems guide our solution design, which systematically addresses each challenge 
through a combination of novel layout schemes, transformation elimination techniques, and global 
optimization strategies as detailed in the following section.

\begin{figure}[!t] 
	\centering
        \includegraphics[width=0.99\linewidth]{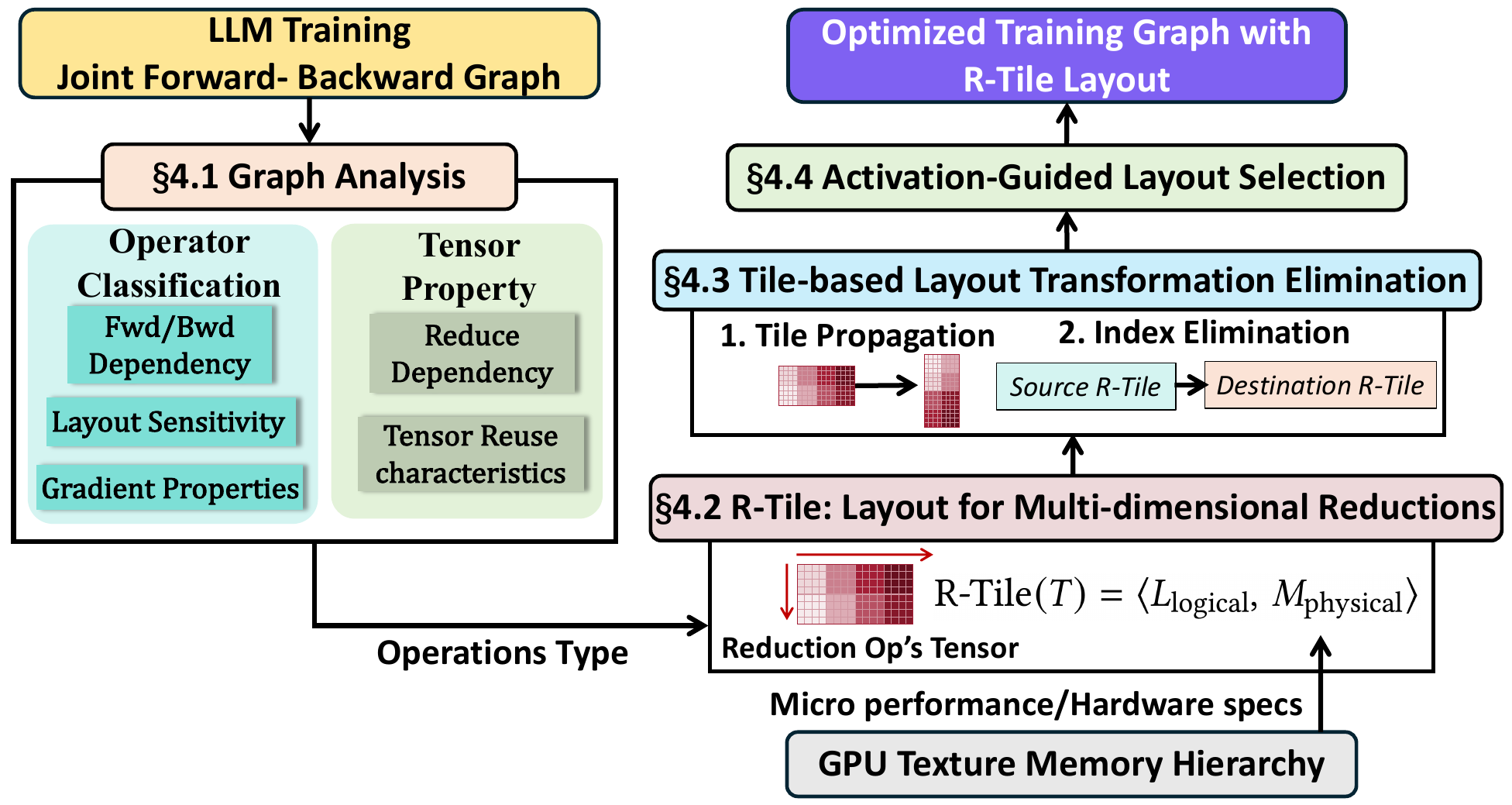}
        \vspace{-5pt}
	\caption{The system overview of \projectname. }
	\label{systemoverview} 
\end{figure}
\vspace{-5pt}

\section{Design of \projectnamenott}\label{sec:design}
This section presents \projectname, a framework that addresses the layout optimization challenges outlined in Sec.~\ref{sec:formalization}. As illustrated in Figure~\ref{systemoverview}, \projectname takes a joint forward-backward training graph as input and produces an optimized graph with cache-efficient layouts through a four-stage pipeline. 
First, \emph{Graph Analysis} (Sec.~\ref{sec:classification}) leverages operator classification to analyze interactions between operators and tensor layouts across forward and backward passes, identifying optimization targets and their associated layout conflicts.
Based on this analysis, \emph{R-Tile} (Sec.~\ref{sec:rtile}) designs unified, texture-aware tile layouts that resolve forward-backward reduction conflicts by exploiting the mobile GPU's 2.5D memory hierarchy ({targeting $Q1$ and $Q4$}). \emph{Tile-based Transformation Elimination} (Sec.~\ref{sec:transform_elim}) then propagates tile structures across adjacent operators and replaces explicit layout transformations with lightweight coordinate remapping ({targeting $Q2$}). Finally, \emph{Global Layout Selection} (Sec.~\ref{sec:global}) globally balances transformation costs against execution efficiency across the entire computation graph ({targeting $Q3$}). The resulting layout plan ensures cache-efficient execution throughout training.


\subsection{Operator Classification}\label{sec:classification}

Optimizing memory layouts for training graphs requires understanding operator interactions 
with tensor layouts across both forward and backward passes. 
Unlike inference with unidirectional data flow, training introduces bidirectional constraints 
where tensors are accessed with divergent patterns during forward computation and gradient calculation. 
To systematically analyze these layout conflicts, 
we develop a classification framework based on two key dimensions:

\begin{itemize}[leftmargin=*,noitemsep,nolistsep]
    \item \textbf{Forward Tensor Dependency}: Whether backward computation requires preserved forward tensors.
    \item \textbf{Access Pattern Consistency}: Whether forward and backward passes exhibit identical memory access patterns.
\end{itemize}

\begin{table}[!t]
\centering
\caption{Operator Classification for LLM training workloads.}
\vspace{-5pt}
\small
\resizebox{\linewidth}{!}{
\begin{tabular}{c|c|c}
\hline
\diagbox{Access Pattern}{Fwd Tensor\\Dependency} & \textbf{No Dependency} & \textbf{Has Dependency} \\
\hline
\textbf{Consistent} & \makecell{\texttt{Add},\\ \texttt{Subtract, Scale}} & \makecell{\texttt{ReLU, Sigmoid,}\\ \texttt{Tanh, Softmax, Dropout}} \\
\hline
\textbf{Divergent} & \makecell{\texttt{Transpose, Permute,}\\ \texttt{Reshape}} & \makecell{\texttt{MatMul, Conv2D,}\\ \texttt{BatchNorm, LayerNorm}} \\
\hline
\end{tabular}
}
\label{tab:training_classification}
\end{table}

These binary dimensions define four operator categories in Table~\ref{tab:training_classification}, each with distinct optimization requirements:

\noindent{\it Forward-Tensor-Independent \& Consistent-Access (FTI-C).}
Operators such as element-wise addition compute gradients purely from upstream information:
\begin{align}
\text{Forward: } & C^{l}_{n\text{-}d} \leftarrow A^{l}_{n\text{-}d} + B^{l}_{n\text{-}d} \nonumber \\
\text{Backward: } & \frac{\partial L}{\partial A}^{l}_{n\text{-}d} \leftarrow \frac{\partial L}{\partial C}^{l}_{n\text{-}d}, \quad \frac{\partial L}{\partial B}^{l}_{n\text{-}d} \leftarrow \frac{\partial L}{\partial C}^{l}_{n\text{-}d} \nonumber
\end{align}
Identical element-wise access patterns impose no layout constraints.

\noindent{\it Forward-Tensor-Dependent \& Consistent-Access (FTD-C).}
Operators like ReLU require preserved forward tensors
 but maintain 
 consistent access:
\[
\frac{\partial L}{\partial X} = \frac{\partial L}{\partial Y} \odot \mathbb{1}_{(X>0)}
\]
Activation storage is required, but any forward-efficient layout remains backward-efficient.

\noindent{\it Forward-Tensor-Independent \& Divergent-Access (FTI-D).}
Transformation operators like transpose and reshape exhibit inverse 
dimension relationships without forward tensor dependencies. 
Forward permutation $(d_1, d_2, d_3) \to (d_2, d_3, d_1)$ requires 
 backward inverse $(d_2, d_3, d_1) \to (d_1, d_2, d_3)$, creating divergent access.

\noindent{\it Forward-Tensor-Dependent \& Divergent-Access (FTD-D).}
Critical operations like matrix multiplication require preserved tensors with conflicting reduction axes:
\[
\frac{\partial L}{\partial W} = X^T \cdot \frac{\partial L}{\partial Y}
\]
Forward reduces along $K$ while backward reduces along $M$, creating fundamental layout conflicts.

This classification reveals increasing optimization complexity and identifies three core challenges:
\begin{itemize}[leftmargin=*,noitemsep,nolistsep]
    \item Resolving forward-backward layout conflicts in FTD-D operators.
    \item Eliminating transformation overhead in FTI-D operators.
    \item Selecting layouts globally while minimizing transformation costs.
\end{itemize}

We address these through R-Tile layouts (Sec.~\ref{sec:rtile}), tile-based transformation elimination (Sec.~\ref{sec:transform_elim}), and activation-guided global selection (Sec.~\ref{sec:global}).

\begin{figure}[!t] 
	\centering
        \vspace{-5pt}
        \includegraphics[width=0.96\linewidth]{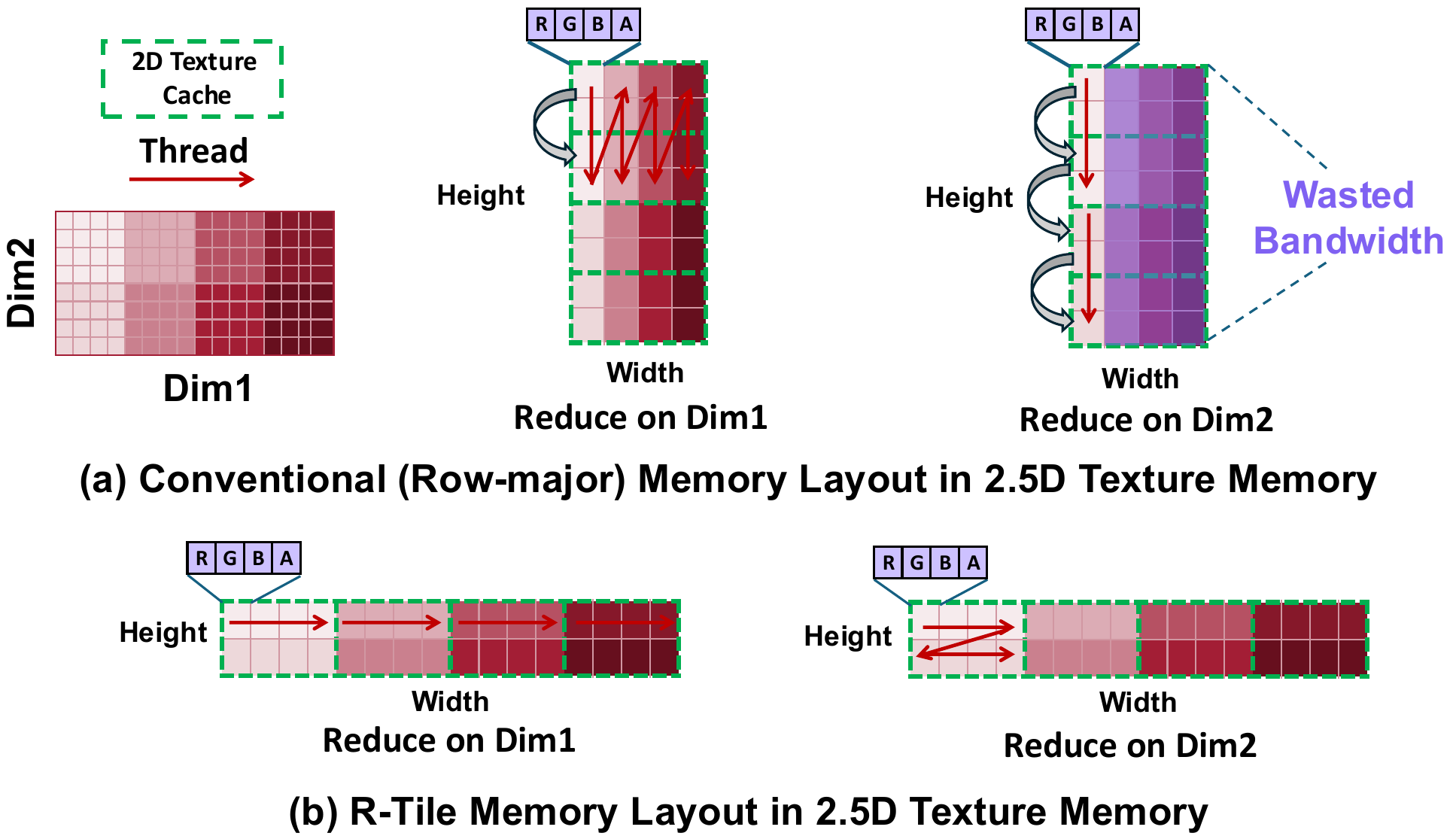}
        \vspace{-10pt}
	\caption{Comparison of conventional memory layout and our R-Tile layout. }
	\label{fig:layoutExample} 
    \vspace{-5pt}
\end{figure}



\subsection{R-Tile: Unified Layout for Multi-dimensional Reductions}\label{sec:rtile}
To resolve the conflicting reduction axes in FTD-D operators identified in our classification, we introduce R-Tile (Reduction-aware Tiling), a unified tensor layout that enables efficient execution of both forward and backward passes without layout switching. The key intuition behind R-Tile is simple: reduction operations over a tile load the same set of elements regardless of the reduction direction---only the aggregation order changes. If we can keep these tile elements physically contiguous for every traversal direction, a single layout suffices for all reductions. R-Tile achieves this idea by exploiting the 2D spatial locality inherent in mobile GPU texture memory, which existing linear layouts fail to leverage. 
We first illustrate why reduction conflicts are difficult to resolve under conventional layouts, then present the tile-invariance observation that motivates R-Tile, followed by the concrete layout design.

The core challenge stems from fundamentally different reduction patterns between passes. Consider matrix multiplication $C_{m,n} = \sum_{k=1}^{K} A_{m,k} \cdot B_{k,n}$: the forward pass reduces along $K$ while the backward pass computes $\frac{\partial L}{\partial B} = A^T \cdot \frac{\partial L}{\partial C}$ by reducing along $M$. Conventional linear layouts optimize for only one reduction direction---row-major provides sequential access along rows but yields strided access along columns, as shown in Figure~\ref{fig:layoutExample}$a$. This conventional layout can only ensure contiguity along a single dimension. In LLM fine-tuning with large tensors, such strided access wastes cache lines that cannot be reused before eviction, severely degrading bandwidth efficiency.

\begin{figure}[!t] 
	\centering
        \includegraphics[width=0.98\linewidth]{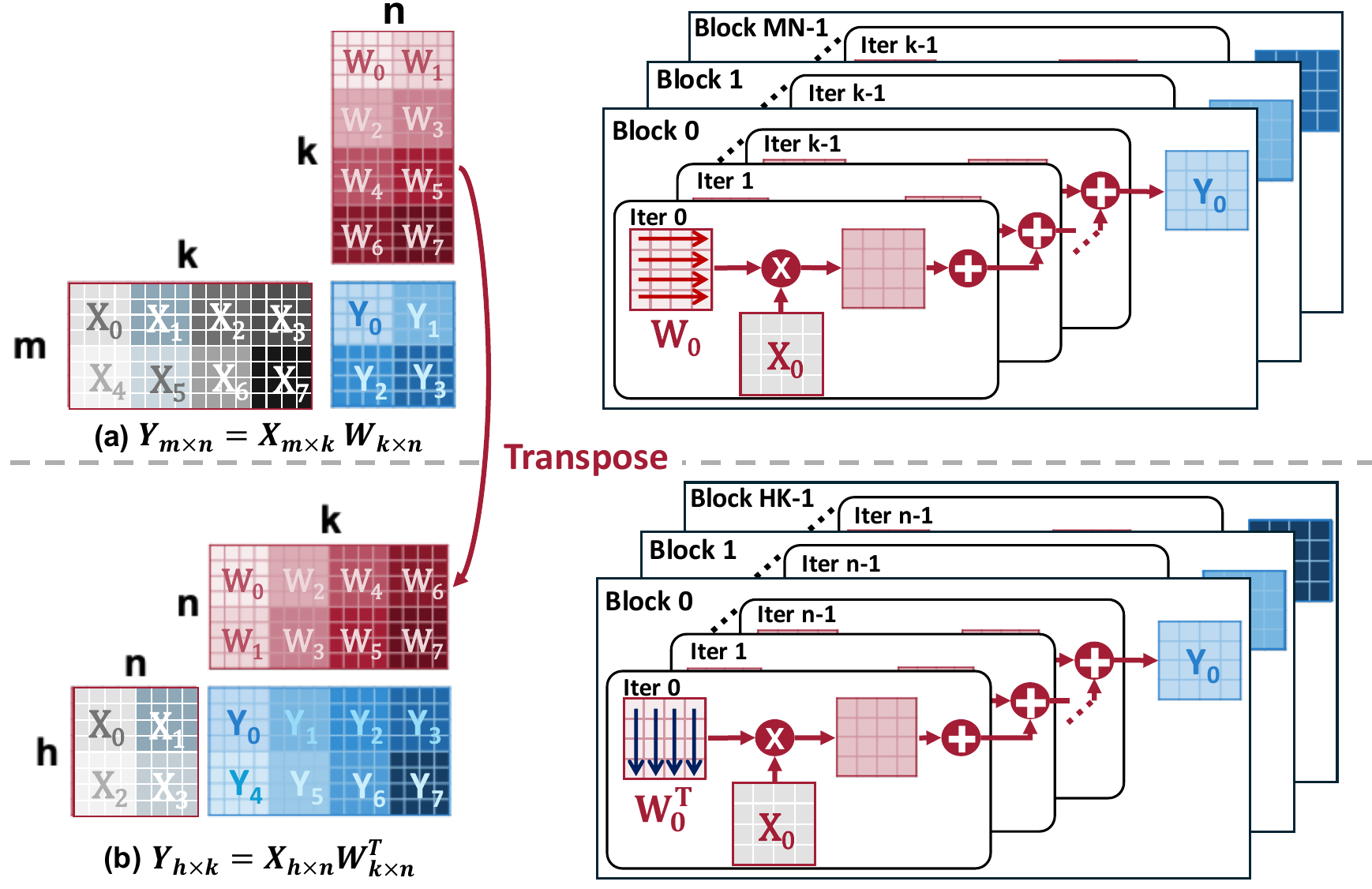}
        \vspace{-5pt}
	\caption{Tile invariance example in matrix multiplication with different reduction axes. Different tiles are shown in varying color shades. (a) Reduce along k-axis; (b) Reduce along n-axis with $W$ transposed to $W^T$. Despite transposition, the same tiles ($W_0, W_1$, ...) are accessed as atomic units: only the traversal direction within each tile differs. }
	\label{exampleReductionUnchanged} 
    \vspace{-5pt}
\end{figure}

Rather than designing separate layouts for each reduction direction, we observe that the tile-level access pattern is inherently invariant to the reduction axis.
Figure~\ref{exampleReductionUnchanged} demonstrates this with {\em MatMul}---both forward (reducing along $K$) and backward (reducing along $N$) passes load identical $(4 \times 4)$ tiles of weight matrix $B$, differing only in aggregation. R-Tile exploits this \emph{tile-invariance} property through two-level organization in 2D texture memory: (1) intra-tile accesses are sequential regardless of the reduction axis, and (2) inter-tile traversal during reductions exhibits stable spatial locality, even when operators reduce along different dimensions.

Given the tensors in reduction conflict FTD-D, R-Tile can statically infer the involved input and output data tiles' shape and layout. R-Tile construction involves two steps: \circled{1} tile shape candidate generation, and \circled{2} physical placement optimization, which are illustrated as below.

\noindent\textbf{1. Tile shape candidate generation.} R-Tile selects tile \texttt{shape} dimensions $(\text{tile}_h, \text{tile}_w)$ covering potential reduction axes, ensuring sequential element access within tiles for any direction. Intuitively, tile shape directly impacts memory access efficiency, cache utilization, and computational throughput \cite{guan2025tmmodel}. 
Existing methods, such as TVM \cite{chen2018tvm}, rely on empirical performance feedback over a large search space, which is time-consuming and impractical for mobile training scenarios. These methods treat the hardware as a black box, ignoring the architectural constraints that inherently invalidate certain configurations. 
Leveraging this insight, R-Tile adopts an architecture-aware pruning strategy that progressively filters candidates through three constraints, ordered by priority: C1 (computation), C2 (memory), and C3 (tensor shape). The complete algorithm is provided in Algorithm~\ref{algfilter} (Supplementary Material); we explain the intuition below.


\textit{C1: Computation Architecture Alignment.} Tile sizes must map efficiently to GPU parallelism. We enforce warp-level alignment—tile dimensions must be multiples of the warp size (32 for ARM Mali, 64 for Qualcomm Adreno)—ensuring that parallel threads fully utilize SIMD execution units without idle lanes. Tiles violating this constraint cannot saturate hardware parallelism.

\textit{C2: Memory Architecture Alignment.} Mobile texture caches have limited capacity and favor specific access patterns. We retain only tiles that: (i) fit within the L1 texture cache, avoiding thrashing, and (ii) align with cache line boundaries (typically 64 bytes), preventing partial line fetches that waste bandwidth. Misaligned tiles cause irregular memory access and degrade throughput.

\textit{C3: Tensor Shape Compatibility.} To minimize padding overhead, we prefer tiles that evenly divide tensor dimensions. When exact divisibility is infeasible, we retain candidates with padding ratios below 10\%, balancing memory efficiency with the architectural requirements from C1 and C2.




Through this progressive constraint enforcement, R-Tile constructs a compact, architecture-aligned candidate space without exhaustive search. The filtered candidates naturally satisfy both hardware feasibility and memory efficiency requirements, providing the foundation for subsequent physical placement optimization.

\begin{figure}[!t]
    \centering
    \includegraphics[width=0.99\linewidth]{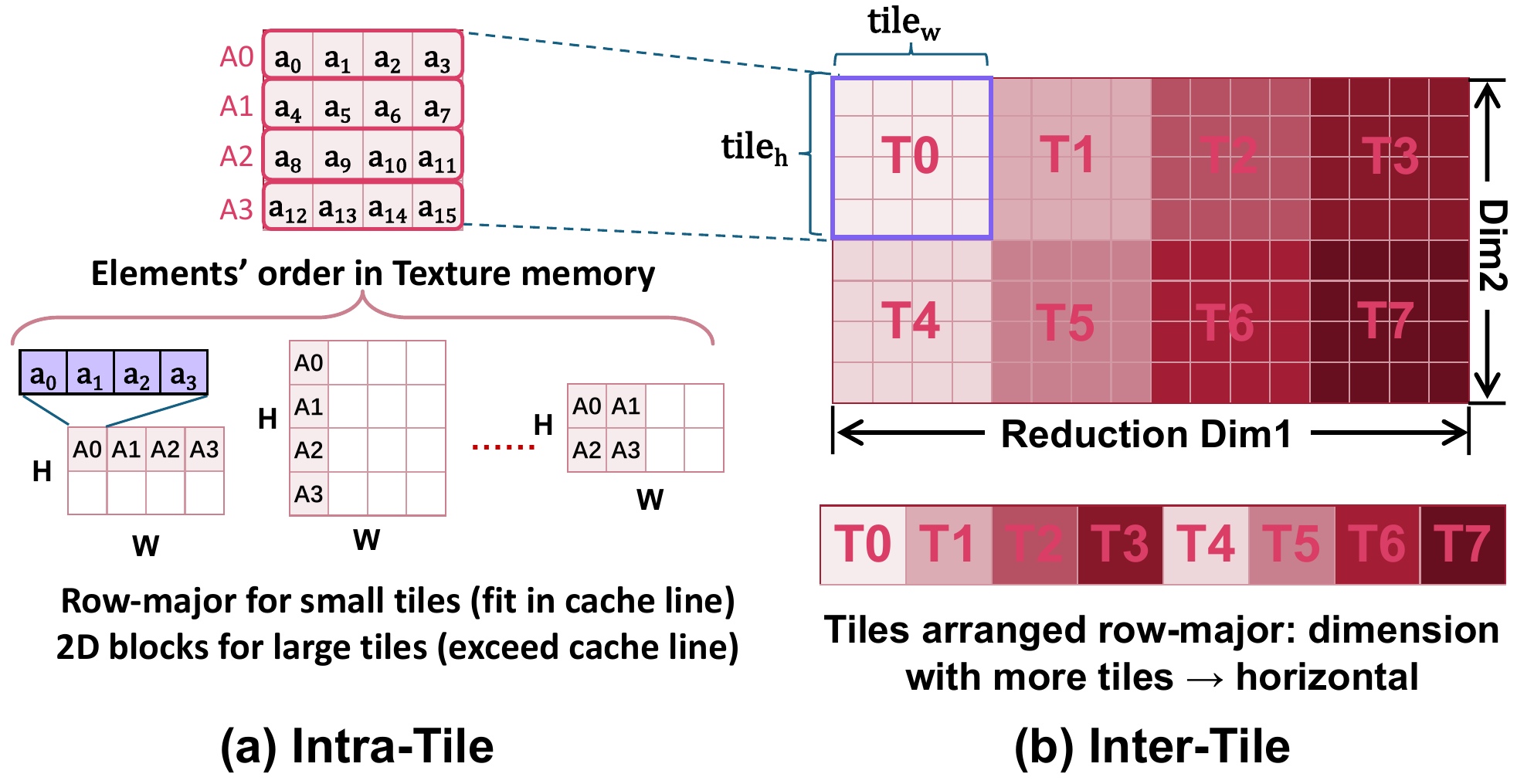}
    \vspace{-5pt}
    \caption{R-Tile layout organization: (a) intra-tile memory packing preserves spatial continuity; 
    (b) inter-tile arrangement in row-major order forms the global tensor layout.}
    \label{fig:rtileexmaple}
    \vspace{-5pt}
\end{figure}

\noindent\textbf{2. Physical Mapping.}
R-Tile maps tensor data into 2D texture memory through two levels, as depicted in Figure \ref{fig:rtileexmaple}. Given tiles with logical dimensions $(\text{tile}_h, \text{tile}_w)$ corresponding to reduction axes:
\textbf{(1) Tile-level mapping (inter-tile):} Arranges tiles in 2D texture space. Tile ordering affects whether consecutive tiles along a reduction axis are spatially adjacent, impacting L2 cache reuse.
\textbf{(2) Element-level mapping (intra-tile):} Arranges the $(\text{tile}_h, \text{tile}_w)$ elements within each tile. Elements can be stored in row-major, column-major, or 2D blocked layouts, directly affecting L1 texture cache efficiency.
Both levels must be co-optimized: inter-tile mapping controls cache reuse across tiles (L2 cache), while intra-tile mapping controls cache efficiency within tiles (L1 texture cache). We address them separately below.


\textbf{Intra-Tile Mapping. } Mobile GPUs fetch texture data in rectangular cache-tile blocks of size ($C_h\times C_w$)(We get these hardware infos based on benchmark experiments like \cite{guan2025tmmodel}). When a warp reads an element, the hardware implicitly loads an entire cache tile, so the efficiency depends on how many of the fetched elements are used before eviction. For an R-Tile with logical dimensions $(\text{tile}_h, \text{tile}_w)$, the number of texture cache tiles it spans is: 
\begin{equation} N{\text{cache}}=\left\lceil \frac{tile_h}{C_h} \right\rceil \left\lceil \frac{tile_w}{C_w} \right\rceil, \end{equation} where minimizing $N{\text{cache}}$ directly improves spatial locality.

We therefore map tile elements adaptively based on tile dimensions relative to cache geometry. When the tile width fits within the hardware cache line width $(\text{tile}_w \le C_w)$, R-Tile stores elements in horizontal strips (row-major order), as any horizontal traversal remains within a single cache line per row. The result in Figure \ref{2DSpatiallocality} confirms that row-major access achieves up to $1.68\times$ throughput when tiles stay within cache line boundaries, maximizing spatial locality for horizontal scans.

However, when $(\text{tile}w > C_w)$, horizontal traversal crosses multiple cache lines, degrading performance. In this regime, R-Tile switches to compact 2D blocks. The specific block dimensions are selected through pre-offline benchmarking, where we profile memory throughput across various tile shapes, including irregular aspect ratios-to identify the configuration that minimizes $N{\text{cache}}$ and aligns best with the target GPU's cache geometry.
This adaptive strategy yields robust intra-tile sequentiality: small tiles use horizontal strips for maximal cache line reuse, while large tiles employ benchmark-selected 2D blocks to minimize active cache tiles under all reduction directions.

\textbf{Inter-Tile Mapping.} Inter-tile mapping determines how tiles are arranged in 2D texture space. The tile ordering directly affects whether consecutive tiles accessed during reductions are spatially adjacent in texture memory, impacting cache locality. Mobile GPU texture caches are organized as 2D blocks for spatial locality, but these blocks are rectangular rather than square: they may span wider in the horizontal direction than the vertical direction. Our texture memory spatial locality experiment in Figute \ref{2DSpatiallocality} confirms this asymmetry: horizontal stride access achieves $1.6\times$ throughput compared with vertical access. Thus, R-Tile applies a simple heuristic: arrange tiles in row-major order, with the dimension requiring more tile traversals as the primary (horizontal) dimension. This keeps consecutive tiles along the heavily traversed dimension spatially adjacent horizontally in texture memory, aligned with the wider cache block extent.

To demonstrate how R-Tile resolves the reduction axis conflict, 
Figure \ref{fig:layoutExample}(b) shows a tensor with shape 
$[\text{Dim2}=8, \text{Dim1}=16]$ using $4\times4$ tiles, under two 
reduction scenarios. In the left panel (Reduce on Dim1), R-Tile maps Dim1 
to the horizontal axis, arranging tiles along the width so the reduction 
traversal proceeds horizontally, fitting entirely within the cache block 
and fully utilizing every fetched cache line. In the right panel 
(Reduce on Dim2), rather than naively placing Dim2 along the vertical 
axis, R-Tile still maximizes horizontal traversal by arranging tiles so 
that the Dim2 reduction path proceeds in a Z-shaped pattern that fits 
entirely within a single 2D texture cache block, with minimal bandwidth 
waste. This is fundamentally enabled by the 2D nature of 
texture memory: by remapping the reduction dimension to the horizontal 
axis, R-Tile exploits the 2D spatial extent of the cache block in a way 
that 1D linear memory cannot support. In contrast, the conventional 
layout in (a) reduces along Dim2 by traversing vertically, crossing 
cache block boundaries and incurring significant wasted bandwidth.

Finally, we select unified R-Tile configurations for tensors consumed by multiple operations with different reduction patterns. Since tensors may be reused by operations reducing along different axes (e.g., a weight tensor used in both forward and backward passes), we choose configurations that maximize overall performance across all consuming operators: 
\begin{equation} \text{R-Tile}^* = \arg\max_{\text{R-Tile}} \sum_{\text{op} \in \text{ReductionOps}} \text{Performance}(\text{R-Tile}, \text{op}) \end{equation} 
where $\text{Performance} = {\text{GFLOPS}} \times \text{MemEfficiency}$. 
$\text{GFLOPS}$ is the measured compute performance, and $\text{MemEfficiency} = \frac{\text{MinimalBytes}}{\text{ActualBytes}}$ quantifies memory traffic efficiency (ratio of theoretical minimum data movement to actual global memory reads). 


\begin{figure}[!t]
    \centering
    \includegraphics[width=0.98\linewidth]{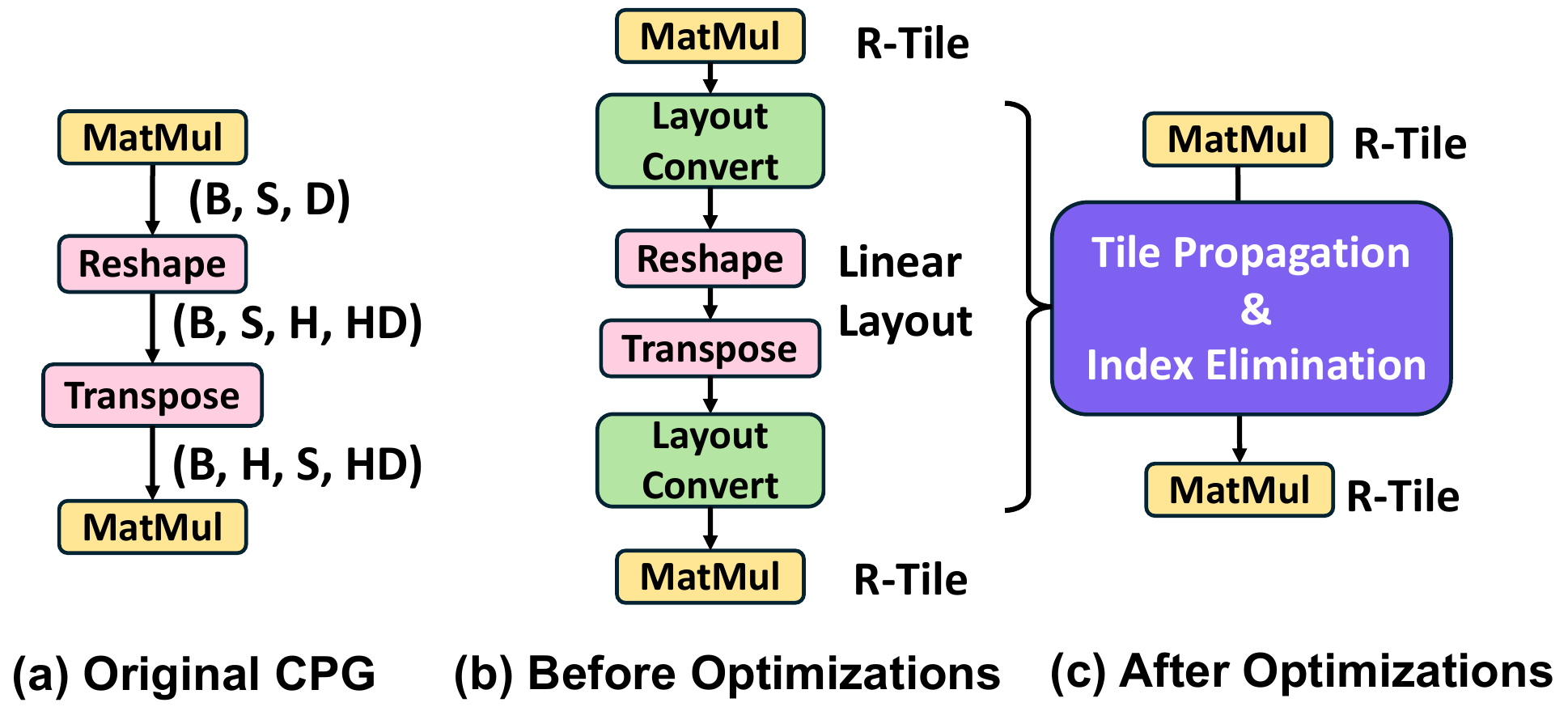}
    \vspace{-5pt}
    \caption{Comparison between elimination-based layout transformation and R-Tile layout in computation graphs. 
    B: batch size, D: hidden dimension, S: sequence length, H: attention heads, HD: per-head dimension.}
    \label{tileoptcompare}
    \vspace{-5pt}
\end{figure}

\subsection{Tile-based Index Transformation}\label{sec:transform_elim}
Layout transformation operators such as Transpose and Reshape (FTI-D 
operators) are pervasive in training graphs and introduce significant 
overhead, particularly under R-Tile layouts where they require costly 
conversions between tiled and linear memory organizations. As shown in 
Figure \ref{tileoptcompare}(b), a subgraph with Reshape and Transpose 
between two MatMul operators requires two explicit Layout Convert nodes 
to switch between tiled and linear representations, adding unnecessary 
data movement overhead.
The key insight behind our approach is that many of these transformations in Transformer-based models only rearrange tiles without altering their internal structure---if a transformation merely reorders tiles while preserving each tile's element layout, we can replace the physical data movement with a lightweight index remapping at tile granularity.
This is far cheaper than element-wise coordinate mapping on the GPU used by conventional approaches \cite{niu2024smartmem,TFlite,jiang2020mnn}, and further enables fusion of adjacent operators by removing the data reorganization barriers between them.
We achieve this through two core techniques: \emph{Tile Propagation} 
analyzes whether a transformation preserves tile structure to verify 
elimination legality, and \emph{Index Transformation Elimination} replaces 
physical data reorganization with simplified index computations when the 
check passes, reducing the subgraph to direct R-Tile-to-R-Tile execution 
as shown in Figure \ref{tileoptcompare}(c).



\subsubsection{Tile Propagation}

This determines whether transformation sequences 
can be expressed as tile-level coordinate remapping without breaking R-Tile structure. 
The key insight is tile invariance: many transformations reorganize outer dimensions 
while leaving reduction dimensions intact, preserving element arrangement within tiles 
and only changing tile positions. 

\begin{figure}[!t] 
    \centering
    \includegraphics[width=0.99\linewidth]{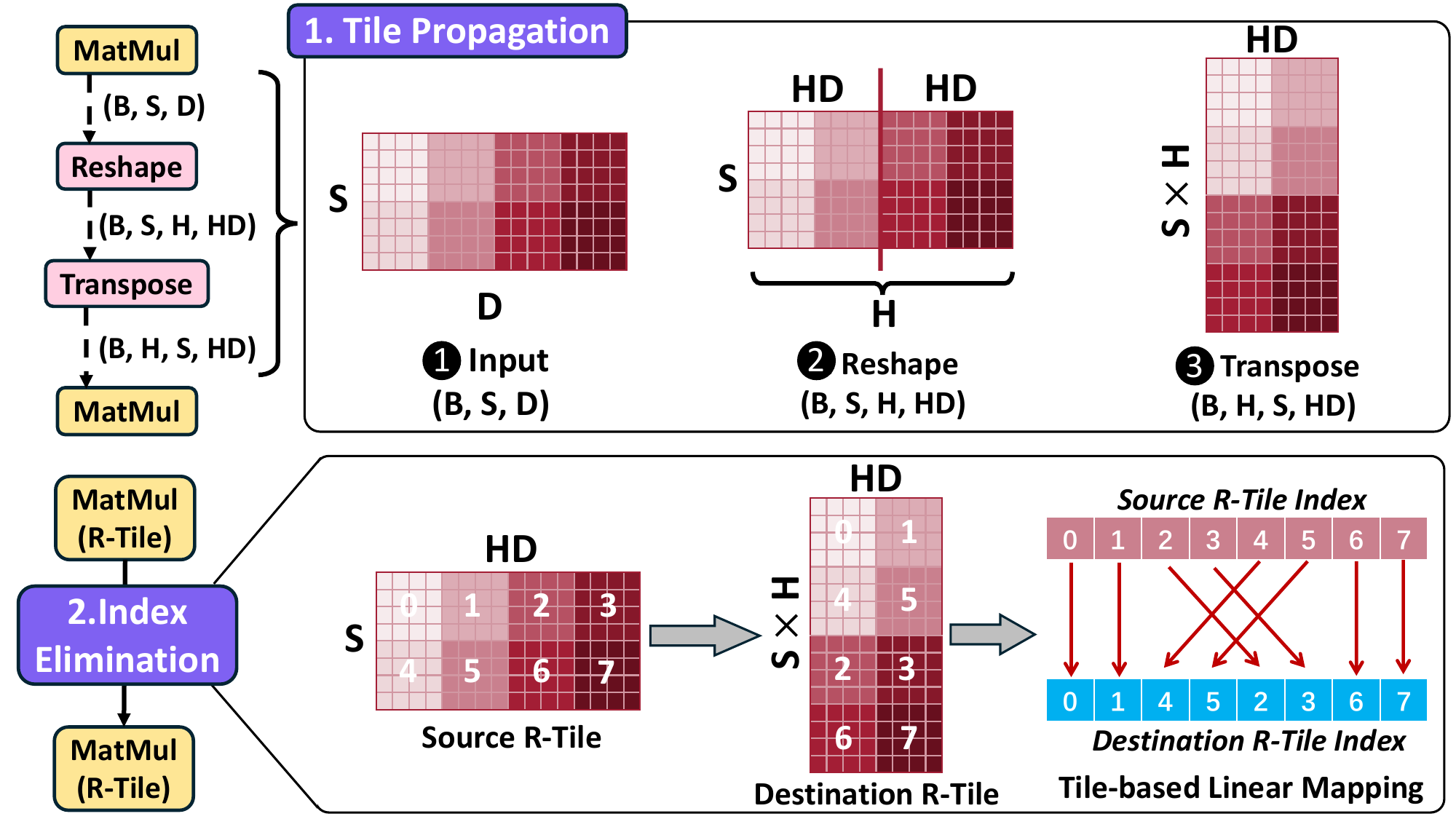}
    \caption{Workflow of the Tile-based Index Transformation. 
    Batch size is set to 1 for illustration. 
    B: batch size, D: hidden dimension, S: sequence length, 
    H: number of attention heads, HD: per-head dimension.}
    \label{TileIndexMethods} 
    \vspace{-15pt}
    
\end{figure}

For example, the top of Figure~\ref{TileIndexMethods} shows a typical reshaping
sequence in multi-head attention, which we use to illustrate tile invariance and
explicitly annotate the role of each dimension at every step. In \circled{1}, the
input tensor \texttt{(B, S, D)} uses the vertical axis for the sequence length $S$
and the horizontal axis for the hidden dimension $D$. In \circled{2}, the tensor is
reshaped to \texttt{(B, S, H, HD)}, where $D$ is factorized into the number of heads
$H$ and the per-head dimension $HD$; here, $H$ appears as the middle horizontal
grouping, while $HD$ remains the innermost axis spanned by each R-Tile. In
\circled{3}, the tensor is transposed to \texttt{(B, H, S, HD)}, swapping the $S$
and $H$ axes while leaving the reduction dimension $HD$ unchanged. Since R-Tile tiles
are defined along $HD$, all three transformations preserve tile atomicity and only
change tile positions, enabling the tile-level index remapping illustrated at the
bottom of the figure.

We verify legality through three dependency checks based on the downstream R-Tile configuration:

\noindent{\it R-Tile Dimension Dependency} leverages the R-Tile determined for reduction operators. 
If matrix multiplication uses R-Tile with ($\text{tile}_w, \text{tile}_h$), 
preceding transformations must preserve these tile dimensions. 
We verify whether dimension manipulations maintain required tile structure.

\noindent{\it Tile Boundary Preservation} ensures subdivided dimensions align with R-Tile boundaries 
for split operations like reshape. 
When reshaping hidden=768 into (heads=12, head\_dim=64) before attention with R-Tile $\text{tile}_w=64$, 
the split must ensure $\text{tile}_w$ is a multiple of head\_dim to maintain alignment. 
$\text{tile}_w=64$ or $\text{tile}_w=8$ with head\_dim=64 both preserve tile boundaries, 
while misalignment breaks tile atomicity.

\noindent{\it Intra-Tile Continuity} verifies that transpose and permute operations 
maintain stride-1 in R-Tile's innermost dimension to preserve contiguous element ordering. 
Loss of stride-1 scatters tile elements across memory.

When all checks pass, tiles propagate through the transformation chain 
with internal structure intact, collapsing the sequence into a single tile coordinate mapping 
compatible with downstream R-Tile layout. 
If verification fails, we attempt recovery through alternative R-Tile configurations 
or adjusted tiled dimensions before falling back to explicit layout conversion. In practice, verification failures occur almost only at steps where the tensor’s element
count changes (e.g., the expansion in grouped-query attention), which temporarily breaks
the previously chosen tile granularity. 
Once the new shape stabilizes, a valid R-Tile
configuration can typically be re-established without fallback.

\subsubsection{Index Transformation Elimination}

Once transformations are verified to preserve tile continuity, 
we eliminate element-wise index computation by replacing transformation operators 
with tile-level index mappings. 
As illustrated in Figure~\ref{TileIndexMethods}, 
we abstract each tile as a single element with row-major numbering, 
enabling tile coordinate translation:
\begin{equation}
T_{\text{dst}}[i] = f(T_{\text{src}}[j])
\end{equation}
where $f$ maps source tile index $j$ to destination tile index $i$. 
This function is precomputed at graph construction time, 
eliminating runtime per-element address calculations.

To further reduce overhead in stacked transformations, 
we apply strength reduction \cite{StrengthReduce} on tile index computation. 
Direct computation using linear representation leads to redundant operations 
involving expensive modulo and division on GPUs. 
We analyze index dependencies and apply mathematical simplification rules -- for example, 
$i \bmod C_a \bmod C_b$ reduces to $i \bmod C_b$ when $C_a \bmod C_b \equiv 0$, 
a common case with tile-aligned dimensions in R-Tile layout.

This becomes particularly effective when multiple transformations are stacked, 
such as reshape followed by transpose in attention mechanisms. 
Instead of computing each transformation separately, 
we compose multiple tile mappings offline:
\begin{equation}
T_{\text{final}} = f_3 \circ f_2 \circ f_1(T_{\text{initial}})
\end{equation}
and apply strength reduction to the composed function. 
This consolidates consecutive transformations into a single simplified tile coordinate function, 
eliminating intermediate transformation kernels entirely.

The resulting simplified index computation enables reduction operators 
to directly access source tiles through composed mapping with minimal overhead, 
converting memory-intensive transformation kernels 
into lightweight coordinate translations while maintaining R-Tile's cache efficiency.

\subsection{Global Layout Optimization}\label{sec:global}

While R-Tile and tile-based transformation elimination address individual operator challenges, training graphs require coordinated layout decisions across operator chains. The core optimization problem involves determining when tensors should adopt R-Tile layouts versus undergo transformations, balancing overhead against efficiency across the entire forward-backward graph. Our approach leverages activation reuse patterns to guide layout propagation. Since tensors (activations) connect producer-consumer operator pairs, their reuse behavior across forward and backward passes determines whether layout conflicts must be resolved. Operators connected through activations with inter-pass reuse impose rigid layout requirements, while those connected through non-reused tensors offer flexibility. By analyzing these producer-consumer relationships, we systematically propagate layouts from conflict-critical operators to flexible ones, minimizing transformations.

\subsubsection{Layout Decision for Operator Combinations}
Our operator classification from Section~\ref{sec:classification} provides the foundation 
for systematic layout propagation across operator combinations. 
When two operators connect through a shared tensor, their layout requirements interact, 
creating patterns that guide our optimization strategy.

\begin{table}[!t]
\centering
\caption{Layout Decision for Operator Pairs in Training Graphs}
\vspace{-10pt}
\small
\resizebox{\linewidth}{!}{
\begin{tabular}{|c|c|c|c|c|}
\hline
\diagbox{Producer}{Consumer} & FTI-C & FTD-C & FTI-D & FTD-D \\
\hline
FTI-C & Fusion & FTD-C & Index Elim & \cellcolor{red!20}FTD-D \\
\hline
FTD-C & FTD-C & FTD-C  & Index Elim & \cellcolor{red!20}FTD-D \\
\hline
FTI-D & Index Elim & Index Elim & Index Elim & \cellcolor{red!20}FTD-D+Elim \\
\hline
FTD-D & \cellcolor{red!20}FTD-D & \cellcolor{red!20}FTD-D & \cellcolor{red!20}FTD-D+Elim & \cellcolor{yellow!30}Greedy Search \\
\hline
\end{tabular}
}

\label{tab:activation_dominance}
\end{table}

Table~\ref{tab:activation_dominance} reveals three key propagation patterns. 
First, FTD-D operators dominate layout choices: when any operator in a pair is FTD-D, 
its R-Tile layout propagates to connected tensors due to rigid requirements 
for resolving forward-backward conflicts. 
Second, FTI-D operators are eliminated through tile-based index transformation, 
requiring no explicit layout decisions. 
Third, FTI-C and FTD-C operators adapt flexibly to propagated layouts, 
as they maintain spatial locality regardless of tile organization.

This systematic propagation identifies FTD-D operators as anchor points 
whose R-Tile layouts flow through flexible operators and absorb transformation operators. 
Layout transformations are inserted only when necessary, such as at graph boundaries 
or for operators incompatible with tile-based representations. 
Crucially, this strategy jointly optimizes both passes—layout decisions based on activations 
requiring inter-pass reuse naturally apply R-Tile choices to both phases, 
while operations without activation reuse inherit layouts from forward counterparts.

\subsubsection{Optimization for Consecutive Layout-Conflicting Operations}

The most challenging scenario occurs when multiple FTD-D operators connect consecutively (yellow cells in Table~\ref{tab:activation_dominance}), such as successive matrix multiplications in attention blocks ($Q \times K^T$ followed by $(QK^T) \times V$). Each operation may prefer different R-Tile configurations based on its specific reduction patterns. Naively inserting layout transformations between every operation introduces prohibitive overhead, yet forcing a single R-Tile configuration across all operations may degrade performance for some.

However, we observe that candidate R-Tile configurations largely overlap across consecutive operations in Transformer models. This stems from transformer architecture similarity: within attention blocks, operations like $Q \times K^T$ and $(QK^T) \times V$ share similar MNK dimensions; within FFN layers, consecutive matmuls also exhibit dimensional regularity. Significant dimensional differences arise primarily between major blocks (e.g., attention vs. FFN), where layout transformations are already amortized over longer computation sequences.

\textit{Resolving R-Tile conflicts.} To handle consecutive FTD-D sequences with potentially conflicting layout preferences, we apply a tensor size-aware greedy search: we identify FTD-D chains, collect their candidate R-Tile configurations, and rank them by input tensor size (larger tensors dominate cost). We then try each configuration in order, attempting to apply it uniformly across the sequence while inserting transformations only where necessary, and select the arrangement with minimum total cost (computation + transformation overhead). For example, in QKV attention where $Q \times K^T$ has larger inputs, we first attempt its preferred R-Tile for both operations, inserting a transformation for $(QK^T) \times V$ only if beneficial. This greedy approach balances optimization quality with tractable compilation time, leveraging the observed candidate overlap to minimize transformation overhead.

\section{Evaluation}\label{sec:evaluation}


\subsection{Implementation} 
Our training system is built on top of MNN (ddd9a61)~\cite{jiang2020mnn}, comprising approximately 9.2K lines of C++ and OpenCL code. We extend MNN with automatic differentiation and operator implementations for all evaluated models to obtain correct training computation graphs. Given the auto-differentiated graph, \projectname parses the complete computation graph into an intermediate representation (IR) and applies a series of graph-level optimization passes for R-Tile layout selection, Tile-based Index Transformation, and Global Layout Optimization. For the optimal layouts identified by R-Tile, we adopt a just-in-time template-based approach to dynamically generate corresponding kernels. For Tile-based Index Transformation, we design a set of rule-based checks that treat tensor layouts as linear mappings, leveraging linear algebra as a unifying abstraction to verify the legality of layout transformations \cite{zhou2026linear}. Expensive index computations introduced by coordinate remapping are eliminated through LLVM optimization passes.

\begin{table}[t!]
        \centering
        \caption{Summary of the 7 evaluated Transformer models, including architecture, modality, parameter size, and fine-tuning hyperparameters.}
        \vspace{-10pt}
        \resizebox{\linewidth}{!}{
                \begin{threeparttable}

                        \begin{tabular}{lccccc}
                                \toprule
                                \textbf{Model}                        & \textbf{Architecture} & \textbf{Input} & \textbf{Params} & \textbf{Batch} & \textbf{Seq./Res./Frames} \\
                                \midrule
                                LLama3.2-1B \cite{dubey2024llama}     & Decoder-only          & Text           & 1.0B            & 1              & 512 tokens                \\
                                Qwen2.5-1.5B \cite{yang2025qwen3}     & Decoder-only          & Text           & 1.5B            & 1              & 512 tokens                \\
                                Gemma2-2B \cite{team2024gemma}        & Decoder-only          & Text           & 2.0B            & 1              & 512 tokens                \\
                                \midrule
                                BERT-Large \cite{devlin2019bert}      & Encoder-only          & Text           & 340M            & 4              & 512 tokens                \\
                                ViT-Large \cite{dosovitskiy2020image} & Encoder-only          & Image          & 304M            & 8              & 3$\times$224$\times$224   \\
                                \midrule
                                Whisper-Large \cite{radford2023robust}\tnote{1}
                                                                      & Enc.--Dec.            & Audio          & 1.55B           & 2              & 300 frames                \\
                                Stable Diffusion v1.5 \cite{Rombach_2022_CVPR}\tnote{2}
                                                                      & Enc.--Dec.            & Image          & 1.4B            & 2              & 512$\times$512            \\
                                \bottomrule
                        \end{tabular}
                        \begin{tablenotes}
                                \footnotesize
                                \item[1] Only the decoder ($\approx$590M parameters)  is fine-tuned.
                                \item[2] Only the UNet ($\approx$860M parameters) is fine-tuned.
                        \end{tablenotes}
                \end{threeparttable}
        }
        \label{modelsummay}
        \vspace{-5pt}
\end{table}

\begin{figure*}[t]
        \centering
        \includegraphics[width=0.98\linewidth]{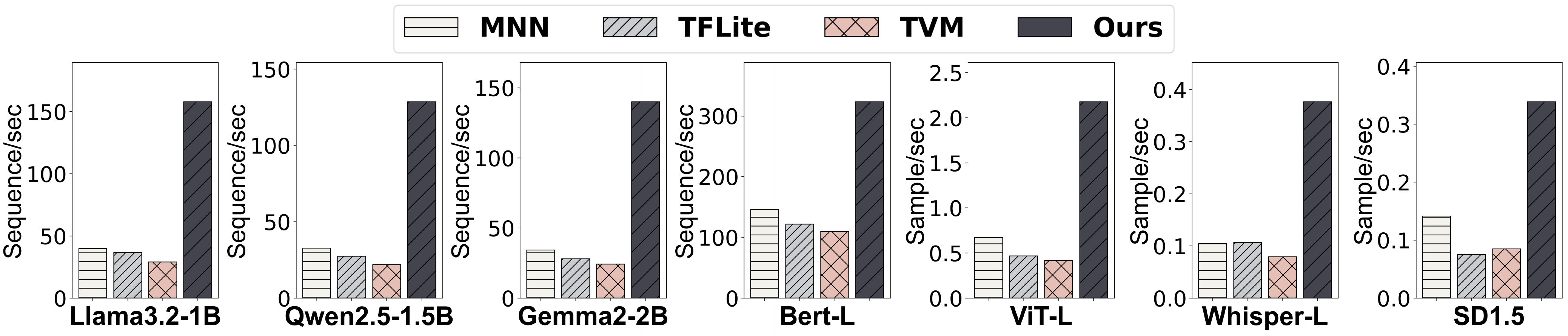}
        \vspace{-10pt}
        \caption{Training speed comparison between other frameworks and \projectname of popular transformer models on OnePlus Ace5Pro (Snapdragon 8Elite).}
        \label{Latencyspeed}
        \vspace{-10pt}
\end{figure*}

\subsection{Experimental Setup}


\compactparagraph{Models.}
We evaluate \projectname using a representative suite of Transformer-based models spanning
the three major architectural families -- encoder-only, decoder-only, and encoder–decoder,
covering both commonly used CV and NLP workloads.
Table~\ref{modelsummay} summarizes the model architectures, modalities, parameter sizes, and fine-tuning settings.
Our encoder-only models include BERT-Large~\cite{devlin2019bert} and ViT-Large~\cite{dosovitskiy2020image};
decoder-only models include Llama3.2-1B~\cite{dubey2024llama}, Qwen2.5-1.5B~\cite{yang2025qwen3}, and Gemma2-2B~\cite{team2024gemma};
and encoder–decoder models include Whisper-Large~\cite{radford2023robust} and Stable Diffusion v1.5~\cite{Rombach_2022_CVPR}.
All experiments use the official pretrained checkpoints.
For all models except ViT, we adopt LoRA-based fine-tuning~\cite{hu2022lora}, inserting LoRA modules into the QKV projection and MLP blocks.
For ViT, we employ adapter-based tuning~\cite{chen2022vision}, which is widely used for vision backbones deployed on resource-constrained devices.
Because \projectname improve the data access and computation efficiency only -- without altering model architectures, tensor precisions, or training objectives -- the fine-tuned accuracy is identical across frameworks;
thus, all comparisons focus on execution efficiency.


\compactparagraph{Baselines.}
We compare \projectname against existing mobile deep learning frameworks: MNN~\cite{MNNBFC}, TVM~\cite{chen2018tvm}, and TensorFlow Lite (TFLite)~\cite{TFlite}.
MNN supports automatic differentiation, whereas TVM and TFLite offer limited training capabilities.
To ensure fairness, we manually construct the joint forward-and-backward computational graphs for each model and feed them to both TVM and TFLite.
This manual graph construction does not affect their runtime performance because their optimizations and operator kernels are architecture-agnostic and operate independently of the graph-generation process.

\begin{figure}[!t]
        \centering
        \includegraphics[width=0.99\linewidth]{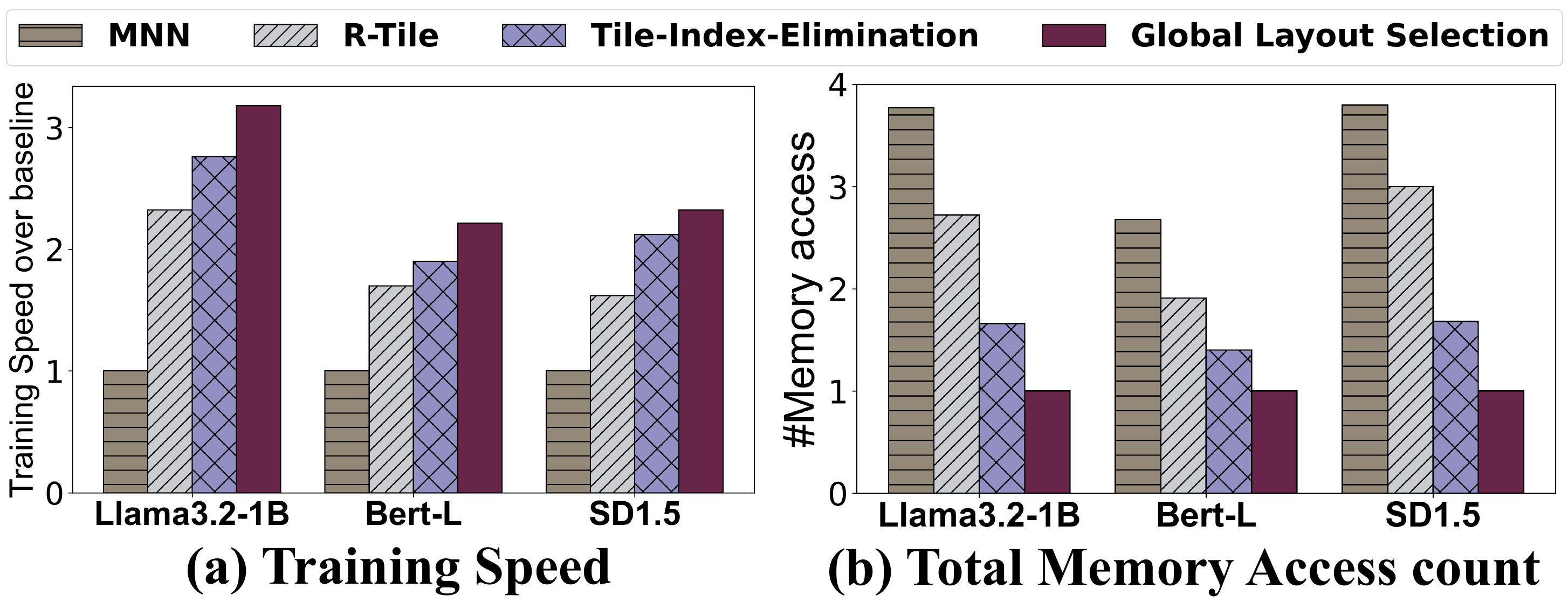}
        \vspace{-10pt}
        \caption{Optimization Breakdown with training speed and memory total access count.}
        \label{ablationstudy}
        \vspace{-10pt}
\end{figure}

\compactparagraph{Target Mobile Devices.}
We deploy \projectname on three representative commercial smartphones equipped with diverse mobile GPUs:
(1) OnePlus Ace5 Pro with {\it Snapdragon 8 Elite}, Adreno 830 GPU, and 16GB RAM;
(2) OnePlus Ace10 Pro with {\it Snapdragon 8 Gen1}, Adreno 730 GPU, and 12GB RAM; and
(3) OnePlus Ace5 Ultra with another hardware vendor ({\it Dimensity 9400+}), Mali Immortalis-G925 GPU, and 16GB RAM.
These devices cover both Qualcomm Adreno and MediaTek Mali GPU families, enabling us to evaluate \projectname across heterogeneous mobile hardware architectures.

\compactparagraph{Metric.}
We measure three primary metrics: training throughput and cache efficiency.
Training throughput is measured as the number of samples processed per second;
and cache efficiency (collected from Snapdragon Profiler~\cite{snapdragon_profiler}) is computed using hardware performance counters as the ratio between
total global memory accesses and cache misses in the mobile GPU.
Because \projectname does not modify model parameters, tensor precisions, or training objectives, model accuracy remains unaffected and is therefore omitted from the reported metrics.


\begin{figure}[!t]
        \centering
        \includegraphics[width=0.99\linewidth]{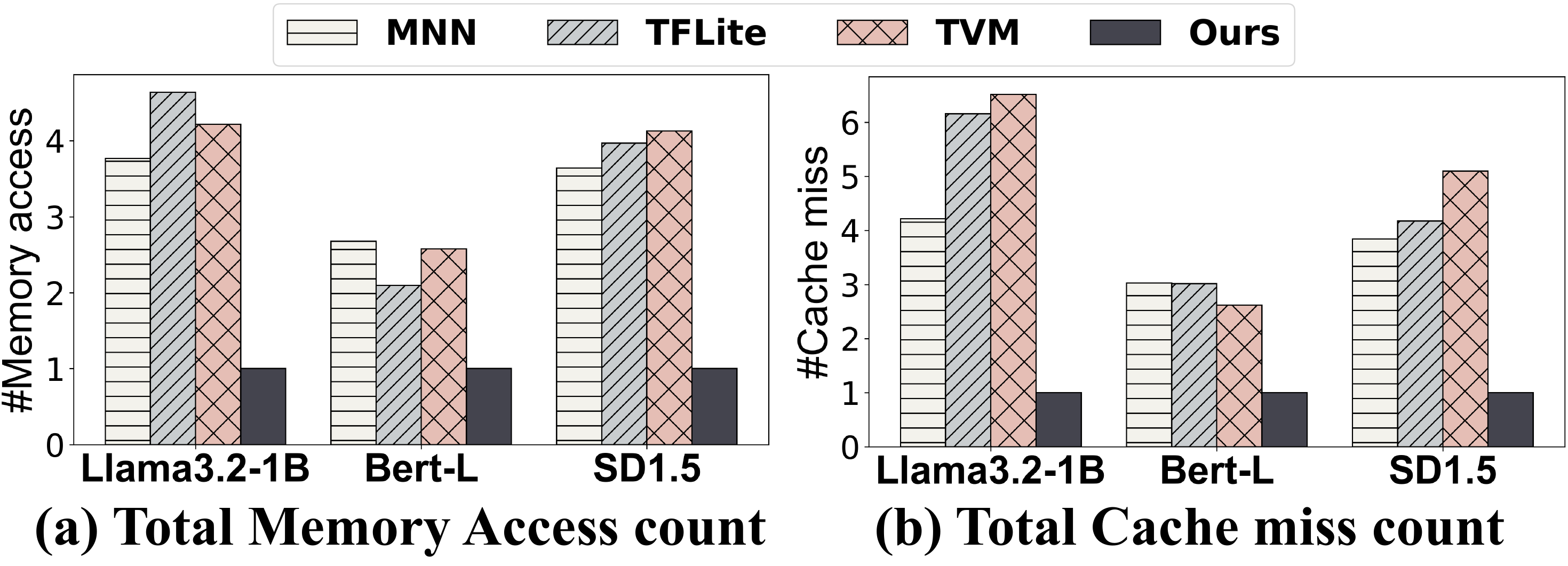}
        \vspace{-10pt}
        \caption{Analysis of Cache Efficiency. All results are normalized by ours.}
        \label{memoryaccesscount}
        \vspace{-20pt}
\end{figure}

\subsection{Overall Performance Evaluation}
\label{sec:overall_performance}

\noindent\textbf{End-to-end latency comparison.}
Figure~\ref{Latencyspeed} reports the end-to-end training latency across
decoder-only, encoder-only, and encoder–decoder Transformer models.
\projectname delivers consistent speedups for all architectures,
achieving $3.9\times$–$4.1\times$ over MNN, $4.3\times$–$4.9\times$ over TFLite,
and $5.4\times$–$5.7\times$ over TVM on decoder-only LLMs.
For encoder-only models, \projectname achieves $2.2\times$–$3.2\times$, $2.6\times$–$4.6\times$,
and $2.9\times$–$5.2\times$ improvements over MNN, TFLite, and TVM, respectively.
Encoder–decoder models observe $2.3\times$–$3.5\times$, $3.5\times$–$4.5\times$,
and $4.0\times$–$4.7\times$ speedups over the same baselines.

It is worth noting that the degree of speedup varies due to structural differences among Transformer architectures. Decoder-only and encoder-decoder models rely on dense QKV interactions, large projection matrices, and frequent reshape and transpose operations; these patterns exacerbate layout conflicts and increase the overhead of kernel-preferred layout conversions in backends such as MNN and TFLite. Stable Diffusion further complicates execution by mixing convolutional and attention layers, intensifying the layout transitions that TFLite handles inefficiently. TVM benefits from operator fusion but lacks a reduction-friendly layout or texture-aware mapping for mobile GPUs, leading to significant performance regressions when reduction or reshaping operators dominate. In contrast, \projectname delivers stable gains across all models by employing a unified R-Tile layout combined with tile-based transformation elimination. This design reduces global memory traffic, maintains operator consistency along critical paths, and avoids the repeated layout conversions that limit existing frameworks. We will provide a more detailed breakdown analysis in the next section.

\subsection{Optimization Breakdown and Analysis}
\label{sec:breakdown}
We next quantitatively evaluate the contribution of each optimization in \projectname. We incrementally enable each component: R-Tile, tile-based index elimination, and global layout selection-creating multiple configurations of \projectname, and compare each configuration against the MNN baseline to measure the performance improvement.

\begin{figure}[!t]
        \centering
        \includegraphics[width=0.98\linewidth]{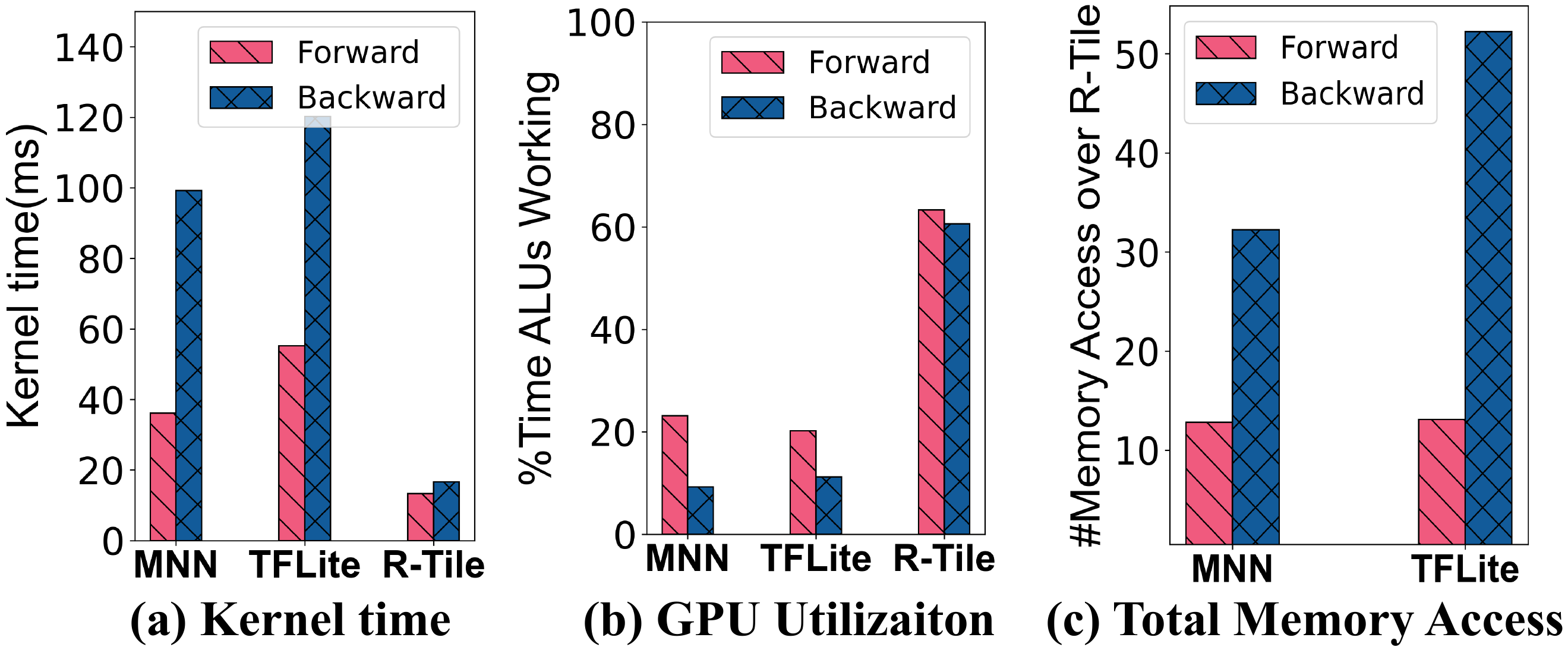}
        \vspace{-5pt}
        \caption{Optimization Breakdown with R-Tile. We benchmark a representative MatMul operator where the forward pass computes $A×B$ with dimensions (M=512, K=2048, N=8192), and the backward pass reuses the same weight matrix by computing $dA = dO×B^T$ with dimensions (M=512, K=8192, N=2048). }
        \label{R-tilAblation}
        \vspace{-10pt}
\end{figure}

Figure~\ref{ablationstudy} summarizes the incremental benefits.
Enabling R-Tile alone yields an average $1.88\times$ speedup across the three models and reduces global-memory accesses by $2.50\times$,
as the unified layout eliminates forward–backward reduction conflicts and generates more contiguous memory-access patterns.
Adding tile-based index elimination provides an additional $1.45\times$ improvement and reduces memory accesses by $1.86\times$ by removing physical layout conversions and exposing additional opportunities for operator fusion.
Incorporating global layout selection contributes a further $1.31\times$ speedup and reduces memory reads by $1.56\times$,
as consistent layout propagation prevents unnecessary R-Tile transitions in memory-intensive regions.
Together, these results demonstrate that the three optimizations provide complementary benefits and collectively enable substantial performance improvements.
In the following sections, we further analyze how individual components influence memory behavior and access locality.

\compactparagraph{Cache efficiency analysis.}
We compare the cache efficiency of \projectname with the baselines.
Figure~\ref{memoryaccesscount} presents the total number of global-memory reads and the corresponding
cache misses for representative models (Llama3.2, BERT-Large, and Stable Diffusion v1.5).
On average, \projectname issues $3.5\times$ fewer global-memory accesses and incurs $4.2\times$ fewer cache misses than MNN, TFLite, and TVM.
Decoder-only LLMs exhibit the largest benefit: \projectname consistently reduces cache misses by more than $4\times$ compared with all baselines.

Large Transformer workloads place heavy pressure on mobile GPU caches due to large projection matrices, multi-head attention operators, and reshape-intensive execution patterns.
Because baseline frameworks apply inconsistent layouts and frequent layout transformations,
they suffer from non-contiguous and low-locality memory accesses, which in turn trigger excessive cache thrashing.
\projectname mitigates this bottleneck by maintaining spatial locality through the R-Tile layout, increasing data reuse,
and reducing global-memory pressure during matrix-heavy and attention-heavy operations.
Section~\ref{sec:breakdown} further analyzes how individual optimizations contribute to these improvements in memory-access efficiency.

\begin{figure}[!t]
        \centering
        \includegraphics[width=0.98\linewidth]{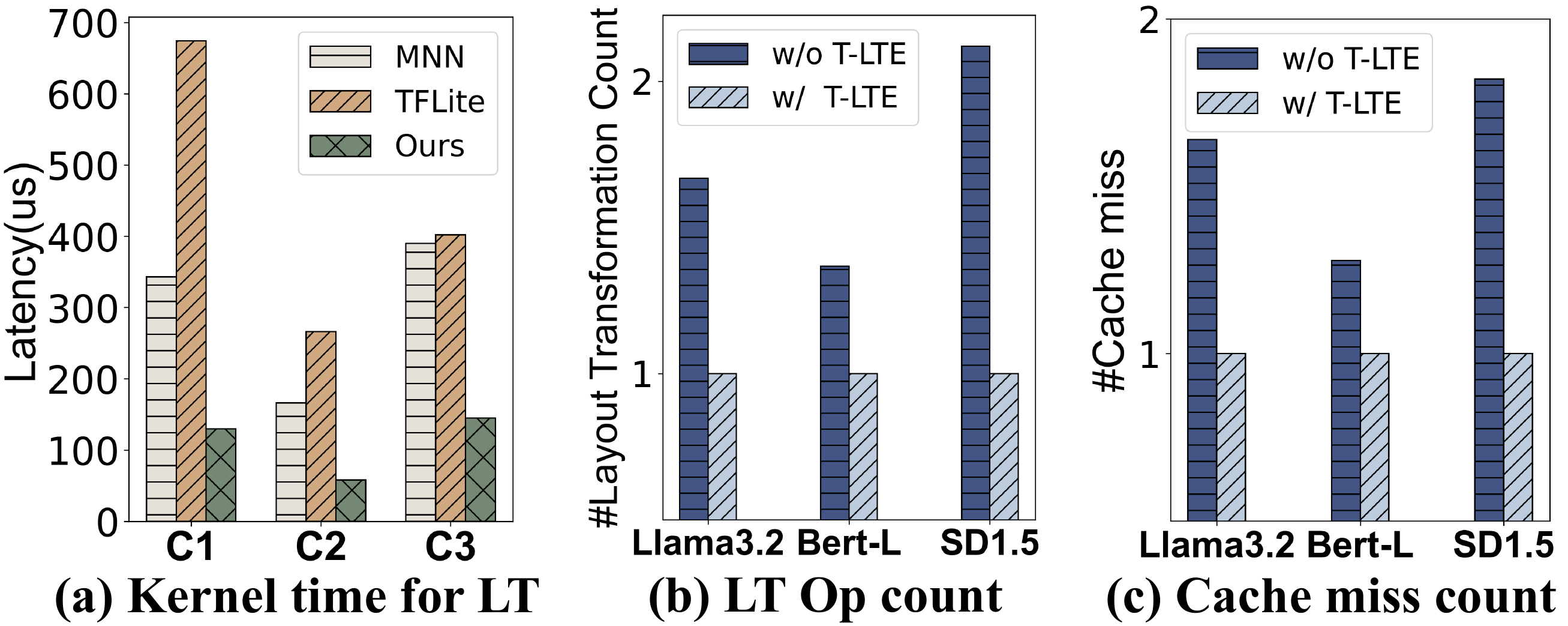}
        \vspace{-5pt}
        \caption{Optimization breakdown with tile-based layout index transformation elimination. LT denotes Layout Transformation. (a) Kernel time for data transformation under different attention configurations \texttt{[sequence length, head\_num, hidden\_dim]}: C1 = (512, 32, 2048), C2 = (512, 16, 1024), and C3 = (512, 8, 1024).}
        \label{analysisTileTransformation}
        \vspace{-10pt}
\end{figure}

\compactparagraph{R-Tile Efficiency.} 
We investigate whether R-Tile's unified layout maintains efficiency across both inference and backward passes. Using identical input settings on LLama3.2-1B, \projectname achieves $1.5\times$/$1.4\times$ speedup over MNN/TFLite in inference and $2.1\times$/$2.4\times$ in the backward pass, demonstrating that a single R-Tile layout effectively serves both computational phases without compromise. 
To understand how R-Tile achieves this, Figure~\ref{R-tilAblation} benchmarks a dominating operator in LLMs (MatMul) in which the forward pass computes $M{=}512$, $K{=}2048$, $N{=}8192$, while the backward pass reuses the same weight matrix in transposed form to compute $M{=}512$, $K{=}8192$, $N{=}2048$. 
This configuration activates reduction operations in both directions, emphasizing the inherent layout conflict between forward and backward passes.
R-Tile achieves an average $3.5\times$ speedup over MNN and $4.6\times$ over TFLite across the two passes, and yields substantially higher ALU utilization ($>60\%$ vs.\ $<21\%$ in MNN/TFLite).
These gains arise from R-Tile's unified, reduction-friendly layout, which preserves spatial locality for both $K$-reduction (forward) and $N$-reduction (backward) without requiring layout switches. 
Existing frameworks, in contrast, use separate kernel-preferred layouts for each pass, which fragments memory access, introduces extra transformations, and reduces computational efficiency.
Note that R-Tile's layout selection is also robust under runtime resource contention: resource sharing affects all R-Tile candidates equally, preserving the relative ranking and optimal selection regardless of background workload.

\compactparagraph{Tile-based index elimination.}
Figure~\ref{analysisTileTransformation} evaluates the impact of tile-based index elimination on layout-related overhead.
In the MHA workload of Figure~\ref{analysisTileTransformation}$(a)$,
the tile-based implementation consistently outperforms both MNN and TFLite,
reducing layout-transformation latency by up to $2.8\times$ and $5.1\times$, respectively.
These improvements are achieved by adjusting layouts at the tile level instead of through element‑wise coordinate mapping, which enhances spatial locality and optimizes texture‑memory bandwidth.
Figure~\ref{analysisTileTransformation}$(b)$ shows that tile-based elimination also reduces the number of explicit layout-transformation
operators by up to $2.1\times$, significantly reducing data reorganization in the computation graph.
This reduces the cache miss rate as well: Figure~\ref{analysisTileTransformation}$(c)$ reports an average $1.5\times$ decrease in cache misses across Llama3.2, BERT-Large, and Stable Diffusion v1.5.
By eliminating reorganization-heavy operators and enabling more aggressive graph fusion,
tile-based index elimination provides both lower transformation overhead and a more locality-friendly execution path.
Regarding coverage, tile-based index remapping successfully handles $>86\%$ of all transformations (
$>95\%$ in LLMs), with the remaining explicit conversions accounting for $<6\%$ of total runtime cost. These fallbacks arise when element sizes differ before and after the transformation, which prevents direct tile-level index remapping.

\compactparagraph{Layout compilation time.}
Compilation time directly impacts usability: excessive overhead makes layout optimization impractical for on-device scenarios where users frequently switch fine-tuning tasks.
Table~\ref{layouttuningtime} reports compilation latency for the Llama3.2-1B training graph.
\projectname reduces tuning time from hundreds or even thousands of seconds -- as observed in MNN, TFLite, and TVM,
to only 6.1 seconds. This reduction stems from two key factors:
R-Tile’s architecture-aware pruning strategy, which filters out infeasible tile candidates based on
texture-memory constraints, and the activation-guided global layout rules,
which dramatically shrink the layout search space by enforcing end-to-end consistency.
Together, these mechanisms make \projectname’s compilation both fast and stable across diverse model architectures.

\begin{table}[!t]
        \centering
        \caption{Tuning time comparison across frameworks. Latency is measured as the total compilation time for the LLama3.2-1B training graph.}
         \vspace{-10pt}
        \resizebox{\linewidth}{!}{
                \begin{tabular}{lccccc}
                        \toprule
                        \textbf{Framework} & \textbf{MNN} & \textbf{TFLite} & \textbf{TVM} & \makecell{\textbf{R-Tile w/o}       \\ \textbf{pruning}} & \textbf{Ours} \\
                        \midrule
                        Tuning Lat. (s)    & 723          & 862             & $>4800$      & 221                           & 6.1 \\
                        \bottomrule
                \end{tabular}
        }
        \label{layouttuningtime}
         \vspace{-5pt}
\end{table}

\begin{figure}[!t]
        \centering
        \includegraphics[width=0.99\linewidth]{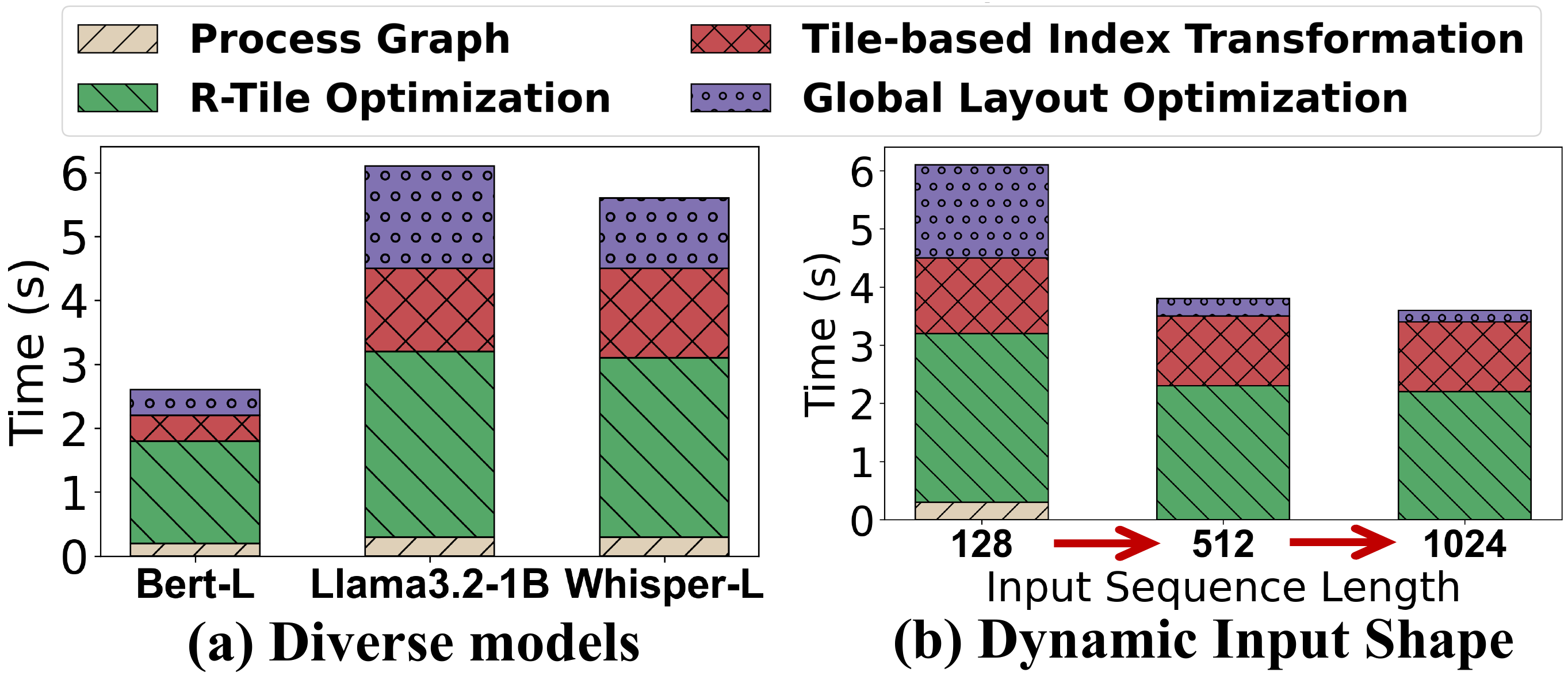}
        \vspace{-10pt}
        \caption{System overhead with different components under various runtime conditions. }
        \label{systemoverhead_dymodel_dyshape}
         \vspace{-10pt}
\end{figure}

\subsection{System Overhead Analysis}
\compactparagraph{Offline Optimization Analysis.} 
\projectname's offline optimization proceeds in four stages: graph processing, R-Tile Optimization, Tile-based Index Transformation, and Global Layout Optimization. 
As shown in Figure~\ref{systemoverhead_dymodel_dyshape}(a), the total overhead across models ranges from 2.6s to 6.1s, all completed before training begins. The overhead remains moderate thanks to Transformer's repetitive architecture limiting unique shape combinations, architecture-aware pruning reducing the search space, lightweight rule checks in Index Transformation, and template-based OpenCL kernel generation requiring no runtime search. For dynamic input shapes, Figure~\ref{systemoverhead_dymodel_dyshape}(b) shows that after initial compilation (6.1s at seq=128 on Llama3.2-1B), shape changes incur lower overhead (3.5s/3.6s at seq=512/1024) because graph processing is skipped and Global Layout Optimization reduces to a lightweight conflict check. Only R-Tile Optimization and Index Transformation re-run when the changed dimension lies on a reduction-conflict axis; otherwise, prior results are reused. Since variation typically occurs along a single dimension, common configurations can be pre-compiled into a lookup table.

\begin{figure}[t]
\centering
\begin{minipage}{0.48\linewidth}
\centering
\includegraphics[width=0.99\linewidth]{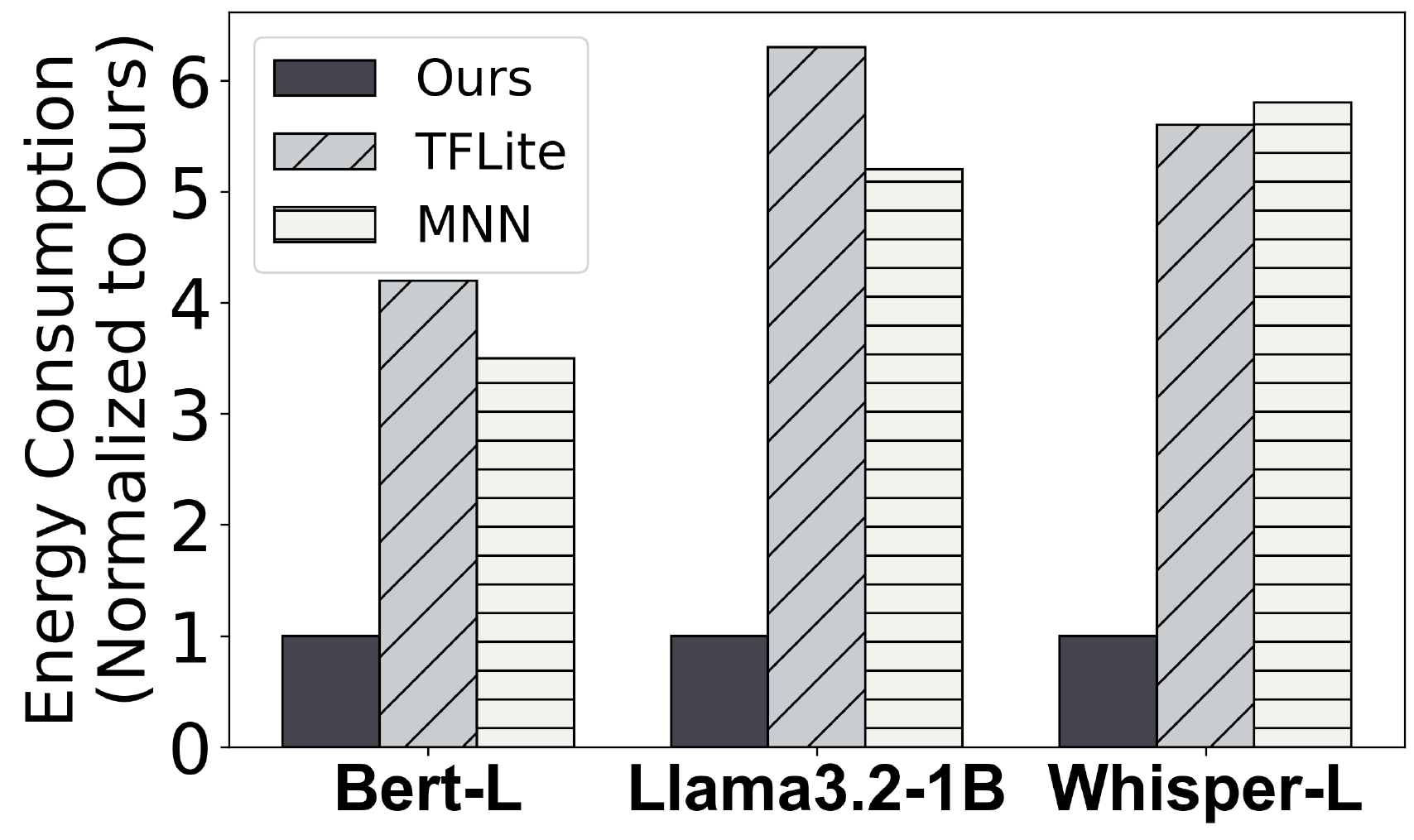}
\vspace{-20pt}
\caption{Energy consumption comparison within 5 rounds of training. Device: Snapdragon 8Elite.  }
\label{fig:energy}
\end{minipage}
\hfill
\begin{minipage}{0.48\linewidth}
\centering
\includegraphics[width=0.99\linewidth]{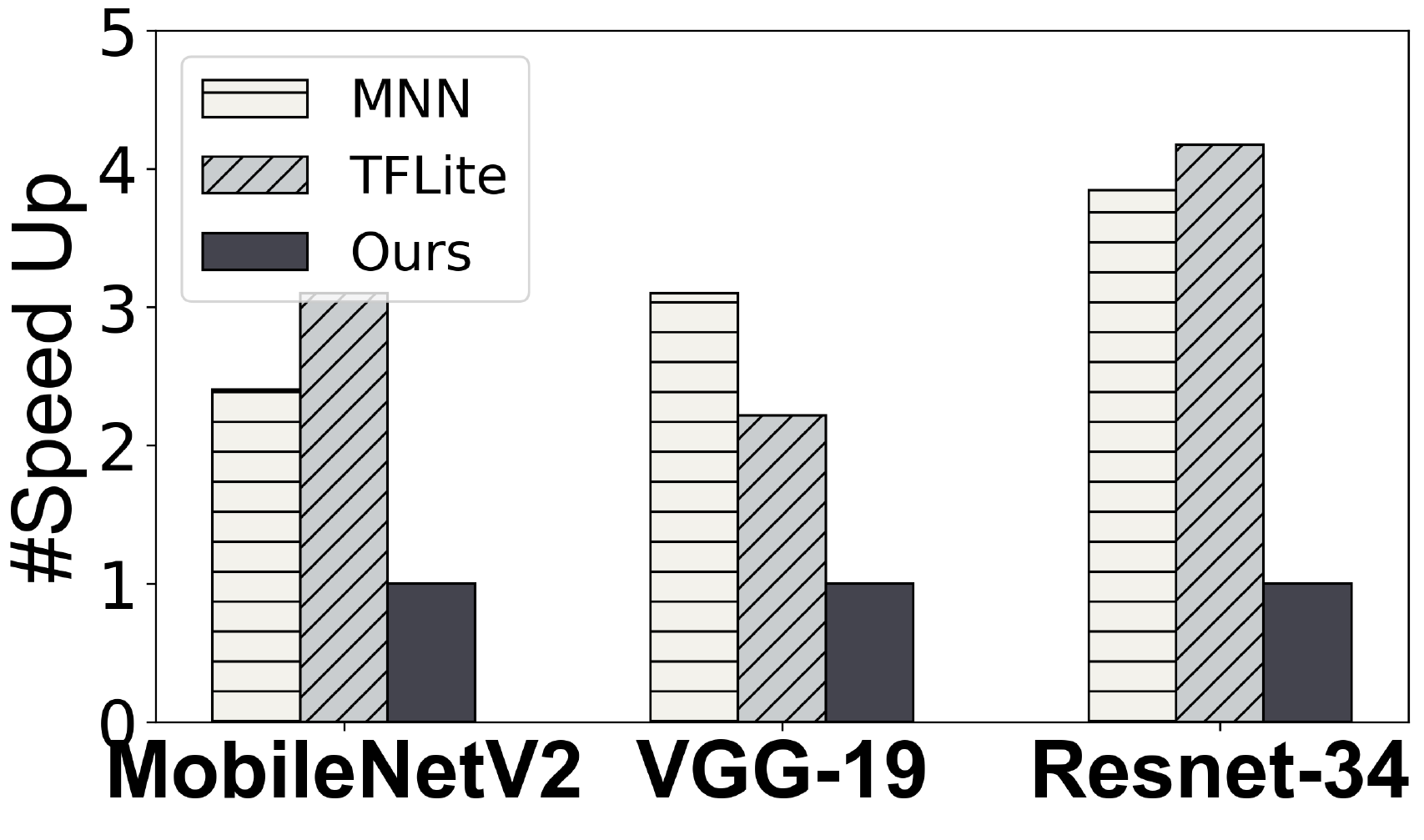}
\vspace{-20pt}
\caption{Portability evaluation on other network architectures. Device: Snapdragon 8Elite.}
\label{fig:othermodel}
\end{minipage}
\end{figure}

\begin{figure}[!t]
        \centering
        \includegraphics[width=0.99\linewidth]{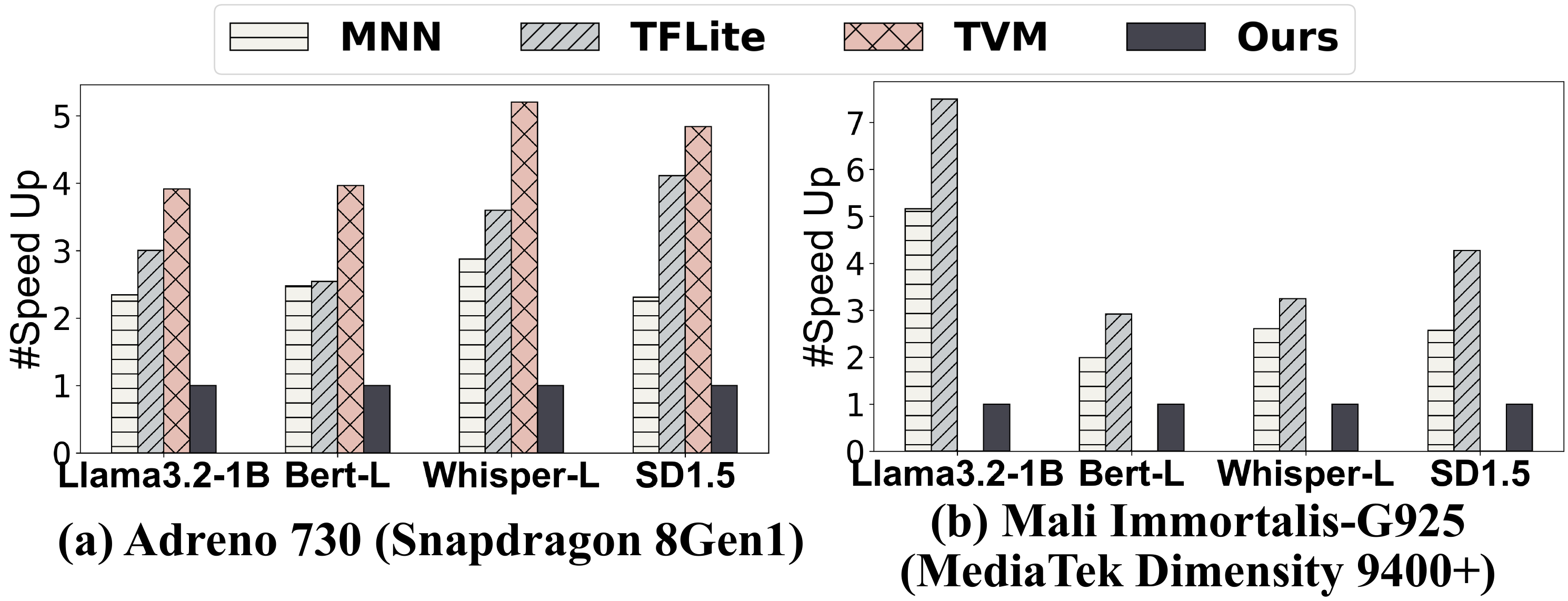}
        \vspace{-10pt}
        \caption{Portability evaluation on Mail GPU and Qualcomm Snapdragon 8Gen1. }
        \label{olderPlatformThroughput}
         \vspace{-5pt}
\end{figure}

\compactparagraph{Effort to Adapt a New Device.} 
Adapting to a new GPU requires a \textit{one-time, GPU-specific, model-agnostic} hardware profiling step that characterizes architectural parameters (e.g., warp size, texture memory L1/L2 cache size and access patterns). These profiles are used to prune the R-Tile search space and can be cached per GPU type and reused across all models. The profiling takes approximately 21s on Snapdragon 8 Elite/8Gen3 (Adreno) and 26s on Mali Immortalis-G925, with no repeated cost at training runtime. New GPU variants require only this profiling step with no additional manual adaptation.

\compactparagraph{Energy Consumption.} We measure energy usage using the Monsoon Power Monitor~\cite{High_voltage_power_moniter} over 5 training rounds across representative models. As shown in Figure~\ref{fig:energy}, \projectname reduces total energy consumption by $3.5$--$6.3\times$ compared to baselines. The energy savings stem from two factors: shorter training time directly reduces active energy expenditure, and fewer cache misses both avoid energy-costly DRAM accesses and eliminate pipeline stalls that would otherwise prolong execution and further increase energy consumption.

\subsection{Portability Evaluation} \label{sec:eva-portability}
\noindent\textbf{Diverse Mobile GPU Platforms:} Figure~\ref{olderPlatformThroughput} shows consistent speedup on MediaTek Dimensity 9400+ and Qualcomm Snapdragon 8Gen1. \projectname generalizes across platforms because R-Tile targets universal texture memory characteristics shared across mobile GPUs rather than GPU-specific features, combined with tile-based transformation elimination that adapts to various architectures.

\noindent\textbf{Other Network Architectures:} To evaluate generality beyond Transformers, we also apply \projectname to CNN training. As shown in Figure~\ref{fig:othermodel}, \projectname achieves $2.41\times$/$3.12\times$ speedup over MNN/TFLite on MobileNetV2 \cite{sandler2018mobilenetv2}, $3.12\times$/$2.22\times$ on VGG-19 \cite{simonyan2014very}, and $3.84\times$/$4.17\times$ on ResNet34 \cite{he2016deep}. These gains confirm that forward-backward layout conflicts also arise in convolution operators, and \projectname's operator classification framework automatically identifies such conflicts and resolves them with optimal layout selection, demonstrating the generality of our approach across diverse network architectures.

\section{Related Work}

\textbf{Optimization for On-device Learning.} Prior work optimizes mobile deep learning through four directions. Memory optimization uses memory pooling \cite{dettmers20218}, selective computation \cite{huang2023elastictrainer,wang2022melon}, activation compression \cite{liu2025daf,tam2024fedhybrid}, and checkpointing \cite{gim2022memory,tam2024fedhybrid} to reduce peak usage. Compilation employs end-to-end compilers like TVM \cite{chen2018tvm} that optimize code via scheduling primitives and operator fusion. Hardware-aware optimization profiles GPU characteristics \cite{liang2022romou} to generate tailored kernels. Layout optimization strategies like SmartMem \cite{niu2024smartmem} optimize tensor layouts and remap indices to eliminate data movement.
These approaches are complementary to \projectname: they reduce memory footprint and improve operator-level efficiency, but still operate on the same forward-backward training graph where layout conflicts between passes persist. \projectname addresses this orthogonal bottleneck by introducing a unified texture-aware layout (R-Tile) that simultaneously optimizes both passes while exploiting mobile GPU texture memory characteristics, rather than avoiding conflicts via fusion or accepting transformation overhead.

\compactparagraph{Efficient Fine-tuning of LLM.} Prior work optimizes LLM fine-tuning through three directions. Parameter-Efficient Fine-Tuning (PEFT) \cite{hu2022lora,houlsby2019parameter,li2021prefix,tian2024hydralora} freezes the base model and updates only adapter modules, prefix embeddings, or low-rank matrices, drastically reducing computational and memory costs. Quantization-Aware Fine-Tuning integrates low-bit quantization with PEFT, such as QLoRA \cite{dettmers2023qlora}, which applies LoRA to quantized models to further compress memory. I/O-Aware Algorithmic Optimization targets memory access efficiency in core operations; FlashAttention \cite{dao2023flashattention} uses compute-memory co-design to minimize data movement. These methods operate at the algorithmic level (reducing parameters, precision, or data movement), whereas \projectname operates at the memory layout level, optimizing how tensors are physically arranged in texture memory. The two are orthogonal because algorithmic methods modify \textit{what} is computed (e.g., fewer parameters via LoRA, lower precision via quantization, fewer memory accesses via FlashAttention), but do not change \textit{how} the resulting tensors are laid out in physical memory—forward-backward reduction conflicts remain in the training graph regardless of these optimizations. \projectname's layout optimization can therefore be applied on top of any such method, addressing the layout-level inefficiency that they leave unresolved.

\section{Discussion}
FBLayout optimizations beyond mobile GPUs. (1) \textit{For desktop-level GPUs: } Desktop GPUs such as NVIDIA \cite{nvhmb} also feature texture memory, making \projectname's R-Tile layout optimization applicable in principle. However, the performance gains may be less pronounced than on mobile platforms, as desktop GPUs benefit from high-bandwidth memory (HBM) \cite{nvhmb} and larger on-chip SRAM, which alleviate the memory access bottlenecks that \projectname targets. Even so, Tile-based Index Transformation remains effective regardless of memory hierarchy, as it eliminates redundant layout conversions in the computation graph. 
(2) \textit{For neural processing units (NPUs): } Current mobile NPUs (e.g., Qualcomm and MediaTek) generally lack support for backpropagation operators and expose limited programmability \cite{npu}. Nevertheless, NPUs rely on specialized tile-based data layouts \cite{hao2025scaling} where bandwidth utilization is critical. \projectname's Tile-based Index Transformation is naturally applicable to such architectures, as it reduces unnecessary data reorganization and bandwidth waste between operators. As NPU programmability and training support mature, \projectname's layout optimization principles can extend to these accelerators.

\section{Conclusion}
This paper introduces \projectname, a novel framework to accelerate LLM fine-tuning on mobile GPUs. \projectname designs a unified R-Tile layout that aligns with mobile GPU texture memory architecture, eliminates redundant layout transformations through tile-based index computation, and propagates efficient configurations globally via activation-guided layout selection. Our extensive experiments with seven Transformer models show $2.2-5.7\times$ speedup compared to MNN, TFLite, and TVM, while substantially reducing memory footprint and improving cache efficiency. 
Future work will combine our layout optimization with other efficiency techniques (e.g., quantization, sparse training) to keep pace with evolving LLM and mobile hardware capabilities.

\begin{acks}
We sincerely appreciate the anonymous shepherd and reviewers for their valuable comments. 
This work is supported in part by the Science and Technology Development Fund of Macau (0107/2024/RIA2, 0061/2025/RIB2), Joint Science and Technology Research Project with Hong Kong and Macau in Key Areas of Nansha District's Science and Technology Plan (EF2024-00180-IOTSC) and the Multi-Year Research Grant of University of Macau (MYRG-GRG2023-00211-IOTSC-UMDF, MYRG-GRG2024-00180-IOTSC).
\end{acks}

\bibliographystyle{ACM-Reference-Format}
\bibliography{sample-base}

\section*{Supplementary Material}

\subsection*{Texture spatial data locality}
We investigate the spatial data locality characteristics of mobile GPU texture caches by examining how different access patterns within a warp affect memory throughput.

\textbf{Experimental configuration.} This experiment uses a single warp with varying thread counts, ranging from 1 to the maximum supported (64 on the tested mobile GPU). Each thread reads different data sections in a streaming manner. The entire data footprint is larger than the L2 cache to rule out the impact of temporal data locality. Each thread reads different data sections in a streaming manner. The entire data footprint is larger than the L2 cache to rule out the impact of temporal data locality. Access pattern design. This experiment is conducted on 4 kernels with representative data access patterns (as shown in Figure~\ref{2DSpatiallocality}(a)). Specifically, they are: 1. Column-major: Each thread reads data in column-major fashion (fixed x-coordinate, iterating y). 2. Row-major: Each thread reads each row sequentially (fixed y-coordinate, iterating x). 3. Block-4: Each thread reads data in a 4-column block manner. 4. Block-8: Each thread reads data in 8-column block manner. Each kernel is executed 10 times, and the average memory throughput is measured.

\textbf{Results Analysis.} Figure \ref{2DSpatiallocality} shows the benchmarking results for 4 access patterns. The x-axis shows the number of active threads in the warp (1 to 64), and the y-axis shows measured memory throughput normalized to column-major (baseline = $1.0\times$). These results unveil two key findings:

1) Regarding 2D spatial locality alignment: Block access patterns (Block-4, Block-8) achieve more consistent performance across thread counts because they align with the texture cache's 2D block structure. Mobile GPU texture caches are organized as rectangular blocks spanning wider horizontally than vertically. Block patterns access data in compact 2D regions that fit within these cache blocks, minimizing cross-block strides. In contrast, row-major access achieves higher peak throughput at low thread counts ($1.68\times$ for 1-8 threads) due to perfect horizontal cache line alignment, but this advantage diminishes as threads collectively span wider regions.

2) Regarding cache line boundary effects: Row-major performance degrades sharply as thread count increases ($1.68\times$ at 8 threads and $1.21\times$ at 64 threads, a 28\% drop), while block patterns remain more stable (Block-8: $1.54\times$ and 1.36\%, only 12\% drop). This occurs because row-major traversal crosses cache line boundaries when threads access data exceeding the cache line width. With many threads scanning different rows simultaneously, the collective working set spans multiple cache lines, causing cache thrashing and contention. Block patterns, by accessing compact 2D regions, keep the working set within fewer cache blocks even as thread count grows.

\subsection*{R-Tile Shape Candidate Generation}
To construct efficient kernels for FTD-D operators on mobile GPUs, we first generate a candidate set of R-Tile shapes that satisfy both hardware constraints and tensor-shape requirements. A naive enumeration of all tile configurations leads to a large design space and includes many candidates that are inefficient on the target architecture. To address this, we introduce a progressive candidate generation procedure, shown in Algorithm~\ref{algfilter}, which begins with an initial tile pool and filters it through three stages. Specifically, the procedure considers: (1) computation architecture alignment, to ensure compatibility with the GPU's execution granularity and ISA constraints; (2) memory architecture alignment, to improve cache efficiency and memory access regularity; and (3) tensor shape compatibility, to reduce padding overhead on irregular tensor dimensions. The final output is a compact candidate set that preserves efficient mappings while avoiding unnecessary search cost.

\begin{algorithm}
\SetAlFnt{\footnotesize}  
\caption{R-Tile Shape Candidate Generation}\label{algfilter}
\SetAlgoLined

\KwIn{Tensors $T$ in FTD-D Operator}
\KwOut{Tile shape candidates $cands$}

$hwInfo \gets \text{GetHardwareInfo}()$\;
$cands \gets \text{InitTileCandidates}(T, FTD, hwInfo)$\;
$cands \gets \text{FilterByComputeArch}(cands, hwInfo)$ \tcp{C1: Warp \& ISA}
$cands \gets \text{FilterByMemoryArch}(cands, hwInfo)$ \tcp{C2: Cache}
$cands \gets \text{FilterByTensorShape}(cands, T.shape)$ \tcp{C3: Padding}
\Return{$cands$}\;

\SetKwProg{Fn}{Function}{:}{}
\Fn{\text{FilterByComputeArch}$(cands, hwInfo)$}{
    $filtered \gets \emptyset$\;
    \ForEach{$cand \in cands$}{
        \If{$\text{IsWarpAligned}(cand) \land \text{IsISACompatible}(cand)$}{
            $filtered.\text{add}(cand)$\;
        }
    }
    \Return{$filtered$}\;
}

\Fn{\text{FilterByMemoryArch}$(cands, hwInfo)$}{
    $filtered \gets \emptyset$\;
    \ForEach{$cand \in cands$}{
        \If{$\text{FitsInCache}(cand) \land \text{IsCacheLineAligned}(cand)$}{
            $filtered.\text{add}(cand)$\;
        }
    }
    \Return{$filtered$}\;
}

\Fn{\text{FilterByTensorShape}$(cands, T.shape)$}{
    $filtered \gets \emptyset$\;
    \ForEach{$cand \in cands$}{
        \If{$\text{ComputePaddingRatio}(cand, T.shape) \leq threshold$}{
            $filtered.\text{add}(cand)$\;
        }
    }
    \Return{$filtered$}\;
}

\end{algorithm}

\textit{Computation Architecture Alignment (C1).} To ensure efficient kernel execution, tile sizes must align with the GPU's parallelism granularity and instruction-level constraints. The \texttt{FilterByComputeArch} function enforces warp-level alignment, ensuring tile sizes are multiples of the warp size (e.g., 32 threads on ARM Mali, 64 on Qualcomm Adreno). Additionally, ISA-specific constraints are applied—for instance, warp-level sync instructions like \texttt{sub\_group\_reduce\_op} require a specific tile dimension operation. This eliminates configurations that cannot efficiently map to hardware execution units.

\textit{Memory Architecture Alignment (C2).} Mobile GPU texture caches exhibit 2D spatial locality with limited capacity. The \texttt{FilterByMemoryArch} function retains only candidates that: (i) fit within the texture cache capacity, and (ii) align with cache line boundaries to avoid partial line fetches. Misalignment causes irregular memory access patterns and cacheline wastage, directly degrading bandwidth utilization.

\textit{Tensor Shape Compatibility (C3).} To minimize padding overhead, \texttt{FilterByTensorShape} favors tile dimensions that evenly divide tensor dimensions. When exact divisibility is infeasible, candidates with padding ratios below a threshold (e.g., 10\%) are retained. This ensures minimal memory waste while maintaining architectural alignment from C1 and C2.

\end{document}